\documentclass[PHME, 2022]{PHMSociety}

\usepackage{microtype} 
\usepackage{setspace} 
\usepackage{amssymb}
\usepackage[T1]{fontenc}
\usepackage{textcomp}

\usepackage{booktabs}    
\usepackage{tabularx}    
\usepackage{multirow}    
\usepackage{makecell}    
\usepackage{caption}     
\usepackage{subcaption}  
\usepackage{float}       
\usepackage{graphicx}    
\usepackage{adjustbox}   
\usepackage[flushleft]{threeparttable} 
\usepackage{stfloats}    

\usepackage{amsmath}  
\usepackage{nccmath}  
\usepackage{amsthm}   
\theoremstyle{definition}

\usepackage{hyperref}
\usepackage{subcaption}
\usepackage{caption}

\usepackage{xcolor, soul} 
\definecolor{darkgreen}{rgb}{0,0.5,0}
\sethlcolor{yellow}      
\usepackage{lipsum}      
\usepackage[normalem]{ulem} 

\usepackage{tikz} 
\usetikzlibrary{automata, arrows, positioning, calc}
\usepackage{tcolorbox} 
\newtcolorbox{mybox1}{colback=red!5!white, colframe=red!75!black}
\newtcolorbox{mybox2}{colback=blue!5!white, colframe=blue!75!black}
\newlength{\RoundedBoxWidth}
\newsavebox{\GrayRoundedBox}
   {\setlength{\RoundedBoxWidth}{#1}
    \begin{lrbox}{\GrayRoundedBox}
       \begin{minipage}{\RoundedBoxWidth}}%
   {   \end{minipage}
    \end{lrbox}
    \begin{center}
    \begin{tikzpicture}%
       \draw node[draw=black, fill=black!10, rounded corners,%
             inner sep=2ex, text width=\RoundedBoxWidth]%
    \end{tikzpicture}
    \end{center}}

\usepackage[nogroupskip, acronym, nonumberlist]{glossaries}
\makeglossaries
\newglossarystyle{mylong}{%
  \setglossarystyle{long}
  ... 
}

\usepackage{paralist}
\newif\ifextended 
\extendedtrue 
\newcommand\smallpar[1]{%
\ifextended%
    \paragraph{#1}%
\else%
    \medskip\noindent\emph{#1}%
\fi%
}

\usepackage{natbib}

\usepackage{etoolbox}
\apptocmd{\thebibliography}{\setlength{\itemsep}{0pt plus 0.3ex}}{}{}

\usepackage{ifthen}    
\usepackage{soulutf8}  
\usepackage{xcolor}    

\newboolean{highlight}
\setboolean{highlight}{false}  

\newcommand{\chl}[1]{%
    \ifthenelse{\boolean{highlight}}{%
        \begingroup%
        \sethlcolor{yellow}
        \hl{#1}%
        \endgroup%
    }{%
        #1%
    }%
}

\graphicspath{{Figures/}} 

\makeglossaries
\begin{document}

\title{Maintenance Strategies for Sewer Pipes with Multi-State Degradation and Deep Reinforcement Learning}

\author{%
Lisandro A. Jimenez-Roa \authorNumber{1},
Thiago D. Simão \authorNumber{2}, 
Zaharah Bukhsh \authorNumber{2}, 
Tiedo Tinga \authorNumber{1}, 
Hajo Molegraaf \authorNumber{3}, 
Nils Jansen \authorNumber{4,5}, and 
Mari\"{e}lle~Stoelinga \authorNumber{1,4}
}

\address{
	\affiliation{{1}}{University of Twente, Enschede, 7522 NB, The Netherlands}{ 
		{\email{\{l.jimenezroa, t.tinga, m.i.a.stoelinga\}}@utwente.nl}
		} 
	\tabularnewline 
	\affiliation{{2}}{Eindhoven University of Technology, Eindhoven, 5612 AE, The Netherlands}{ 
		{\email{\{t.simao@tue.nl, z.bukhsh\}}@tue.nl}
		} 
	\tabularnewline 
	\affiliation{{3}}{Rolsch Assetmanagement, Enschede, 7521 AG, The Netherlands.}{ 
		{\email{hajo.molegraaf@rolsch.nl}}
		} 
    \tabularnewline 
	\affiliation{{4}}{Radboud University, Nijmegen, 6525 XZ, The Netherlands.}{ 
		{\email{n.jansen@science.ru.nl}}\\ 
		} 
  \tabularnewline 
	\affiliation{5}{Ruhr-University Bochum, Bochum, 44801, Germany}
}

\maketitle
\pagestyle{fancy}
\thispagestyle{plain}

\phmLicenseFootnote{Lisandro A. Jimenez-Roa}

\begin{abstract}
Large-scale infrastructure systems are crucial for societal welfare, and their effective management requires strategic forecasting and intervention methods that account for various complexities. Our study addresses two challenges within the Prognostics and Health Management (PHM) framework applied to sewer assets: modeling pipe degradation across severity levels and developing effective maintenance policies. We employ Multi-State Degradation Models (MSDM) to represent the stochastic degradation process in sewer pipes and use Deep Reinforcement Learning (DRL) to devise maintenance strategies. A case study of a Dutch sewer network exemplifies our methodology. Our findings demonstrate the model's effectiveness in generating intelligent, cost-saving maintenance strategies that surpass heuristics. It adapts its management strategy based on the pipe's age, opting for a passive approach for newer pipes and transitioning to active strategies for older ones to prevent failures and reduce costs. This research highlights DRL's potential in optimizing maintenance policies. Future research will aim improve the model by incorporating partial observability, exploring various reinforcement learning algorithms, and extending this methodology to comprehensive infrastructure management.
\end{abstract}

\vspace{-0.1cm}
\newacronym{DRL}{DRL}{Deep Reinforcement Learning}
\newacronym{MSDM}{MSDM}{Multi-State Degradation Model}
\newacronym{MPO}{MPO}{Maintenance Policy Optmization}
\newacronym{PHM}{PHM}{Prognostics and Health Management}
\newacronym{MDP}{MDP}{Markov Decision Process}
\newacronym{DTMC}{DTMC}{Discrete-time Markov chain}
\newacronym{PPO}{PPO}{Proximal Policy Optimization}
\newacronym{IHTMC}{IHTMC}{Inhomogeneous Time Markov Chain}
\newacronym{DQN}{DQN}{Deep Q-Network}
\newacronym{RL}{RL}{Reinforcement Learning}

\printglossary[type=\acronymtype,title=Abbreviations]

\vspace{-0.1cm}
\section*{Abbreviations}
\begin{description}
   \item[DRL] Deep Reinforcement Learning
   \item[IHTMC] Inhomogeneous Time Markov Chain
   \item[MDP] Markov Decision Process
   \item[MPO] Maintenance Policy Optimization
   \item[MSDM] Multi-State Degradation Model
   \item[PPO] Proximal Policy Optimization
   \item[RL] Reinforcement Learning
\end{description}

\section{Introduction}
\label{sec:introduction}

Sewer network systems, crucial for public health, population well-being, and environmental protection, require maintenance to ensure their reliability and availability \citep{cardoso2016sewer}. This maintenance is challenged by limited budgets, environmental changes, aging infrastructure, and hard-to-predict  system deterioration \citep{tscheikner2019sewer}.

Optimizing maintenance policies for sewer networks requires methodologies that can efficiently explore a broad solution space while adapting to the system's dynamic constraints and complexities. \gls{MPO} addresses these needs by developing and analyzing mathematical models to derive maintenance strategies \citep{de2020review} that reduce maintenance costs, extend asset life, maximize availability, and ensure workplace safety \citep{ogunfowora2023reinforcement}.

This research explores the potential of \gls{DRL} for \acrshort{MPO} of sewer networks\chl{, first focusing on a component-level (i.e., pipe-level) analysis}. \acrshort{DRL} is a framework that merges neural network representation learning capabilities with \gls{RL}, a branch of machine learning known for its effectiveness in sequential decision-making problems. \acrshort{RL} is increasingly recognized for its role in developing cost-effective policies in \acrshort{MPO} across diverse domains such as transportation, manufacturing, civil infrastructure and energy systems. It is emerging as a prominent paradigm in the search for optimal maintenance policies \citep{marugan2023applications}. 

This paper aims to achieve two primary objectives: first, to present a comprehensive model for pipe-level \acrshort{MPO} analysis facilitated by \acrshort{DRL}, considering degradation over the pipe length and employing inhomogeneous-time Markov chain models to simulate the nonlinear stochastic behavior associated with sewer pipe degradation; second, to assess the efficacy of the model's policy through a case study of a large-scale sewer network in the Netherlands, comparing it with heuristics, including condition-based, scheduled, and reactive maintenance.

We acknowledge as limitations in our approach the focus on \emph{fully observable} state spaces, which means that inspection actions are not necessary, and our analysis is at the \emph{component-level}. Future research will aim to broaden this scope to include partially observable state spaces and system-level analysis.

\vspace{-15pt}
\paragraph{Contributions.} This work's primary contributions include:
\begin{compactenum}[(i)]
    \item We propose a framework to carry out maintenance policy optimization for sewer pipes considering the deterioration along the pipe length. This framework integrates Multi-State Degradation Models (MSDMs) and Deep Reinforcement Learning (DRL).
    \item Our framework introduces a novel approach by encoding the prediction of the MSDM into the state space, aiming to harness prognostics that describe the degradation pattern of sewer pipes.
    \item We demonstrate that DRL has the potential to devise intelligent strategic maintenance strategies adaptable to various conditions, such as pipe age.
    \item We provide our framework in Python and all data used in this study at \href{https://zenodo.org/records/11258904}{zenodo.org/records/11258904}.
\end{compactenum}

\vspace{-15pt}
\paragraph{Paper outline.} Section \ref{sec:background} presents the technical background. Section \ref{sec:methodology} outlines our research methodology. Section \ref{sec:msdm} formulates the MSDM. Section \ref{sec:model} details the framework for maintenance policy optimization via DRL. Section \ref{sec:experimental_setup} presents our experimental setup. Section \ref{sec:results} analyzes the results. Section \ref{sec:discussion_conclusions} discusses findings, concludes, and suggests future research.

\vspace{-15pt}
\smallpar{Related work.} In the past two decades, the need for integral sewer asset management has become evident \citep{abraham1998optimization}, emphasizing the necessity to understand the mechanisms of deterioration and develop predictive models for proactive and strategic sewer maintenance \citep{fenner2000approaches}. 
Sewer asset management encompasses maintenance, rehabilitation, and inspection and has been investigated through various methodologies, including risk-based strategies \citep{lee2021risk}, multi-objective optimization \citep{Mohamed2019multi}, Markov Decision Processes \citep{wirahadikusumah2003application}, considering the structure of the sewer network \citep{qasem2021gis}, machine learning applications \citep{montserrat2015using,caradot2018practical,laakso2019sewer,hernandez2021optimizing}, and decision support frameworks \citep{taillandier2020decision, khurelbaatar2021data,ramos2022comprehensive,assaf2023optimal}.

The integration of RL into sewer asset management is largely unexplored, with existing research mainly concentrating on \emph{real-time control} for smart infrastructure, adapting to environmental changes such as storms. \cite{mullapudi2020deep} uses \acrshort{DRL} for controlling storm water system valves through simulation of varied storm scenarios. 
\cite{yin2023optimal} employ RL for \emph{near real-time} control to minimize sewer overflows. Meanwhile, \cite{zhang2023towards} and \cite{tian2022combined} both examine improving the robustness of urban drainage systems, the former through decentralized \emph{multi-agent RL} and the latter through \emph{Multi-RL}, with \cite{tian2024improving} further improving the model \emph{interpretability} using \acrshort{DRL}. Furthermore, \cite{kerkkamp2022grouping} investigates the sewer network \acrshort{MPO} by combining \acrshort{DRL} with Graphical Neural Networks to optimize maintenance actions grouping. \cite{jeung2023data} proposes a \acrshort{DRL}-based \emph{data assimilation} methodology to enhance storm water and water quality simulation accuracy by integrating observational data with simulation outcomes.
\section{Technical background}\label{sec:background}

\subsection{Multi-state degradation model for sewer pipes}\label{sec:msdm_background}

The modeling of sewer pipe network degradation has been explored through various methodologies, including physics-based, machine learning, and probabilistic models. For comprehensive discussions on this topic, the reader is directed to \cite{ana2010modeling,hawari2017simulation,malek2019sewer, saddiqi2023smart,zeng2023progress}.

We adopt a probabilistic approach employing \acrfullpl{IHTMC} to model the multi-state degradation of sewer pipes. This choice is motivated by the \acrshort{IHTMC}'s capability to better capture the degradation of long-lived assets such as sewer systems as a non-linear stochastic process, characterized by age-dependent transition probabilities between degradation states \citep{jimenez2024comparison}.

\vspace{-15pt}
\smallpar{\acrfullpl{IHTMC}.} An \acrshort{IHTMC} is a stochastic process $\{(X_t)\}_{t \geq 0}$, where $t \in [0, \infty)$ is continues and models \emph{time}. The \acrshort{IHTMC} is defined as a tuple $M = \langle \Omega, S^0, Q(t) \rangle$, where $\Omega$ is a set of $K$ finite states indicating the \emph{state space}, $S_k^0$ is an \emph{initial-state distribution} on $\Omega$ where $\sum_{k \in \Omega} S_k^0 = 1$, and \(Q(t): \Omega \times \Omega \to \mathbb{R}\) is a \emph{time-dependent transition rate matrix}, with entries $q_{ij}(t)$ for $i,j \in \Omega$ and $i \neq j$, representing the rate of transitioning from state $i$ to state $j$ at time $t$. The diagonal entries $q_{ii}(t)$ are defined such that the sum of each row in $Q(t)$ is zero, ensuring that the \emph{outflow} from any state is equal to the sum of the \emph{inflows} into other states. $Q(t)$ may be parameterized by hazard rates $\lambda(t|\theta)$ derived from the ratio $f(t|\theta)$ and $S(t|\theta)$, being respectively a \emph{probability density function} and a \emph{survival function}, where $\theta$ corresponds to the function hyper-parameters. The evolution over time of the \acrshort{IHTMC} is governed by the \emph{Forward Kolmogorov} equation:

\vspace{-10pt}
\begin{equation}
\frac{\partial P_{ij}(t,\tau)}{\partial t} = \sum_{k \in S} P_{ik}(t,\tau)Q_{kj}(t)
\label{eq:forward_kolmogorov}
\end{equation}
\vspace{-10pt}

Here, \(P_{ij}(t, \tau): \Omega \times \Omega \to [0,1]\) is a continuous and differentiable function known as the \emph{transition probability matrix}, indicating the probability of transitioning from state \(i\) to state \(j\) in the time interval \(t\) to \(\tau\), where \(\tau > t\). From Eq.~(\ref{eq:forward_kolmogorov}) one can obtain the \emph{master equation of the Markov chain}, which models the flow of probabilities between states by including inflow and outflow terms:

\vspace{-10pt}
\begin{equation}
\frac{\partial S_{k}(t)}{\partial t} = \sum_{i \in \Omega, i \neq k  } S_{i}(t) Q_{ik}(t) - S_k(t) \Big( \sum_{j \in \Omega, j \neq k}  Q_{kj}(t) \Big) 
\label{eq:master_MC_equation}
\end{equation}
\vspace{-10pt}

Here, \(S_k(t)\) is the probability of being in state \(k \in \Omega\) at time \(t\), the term \(\sum_{\substack{j \in \Omega , j \neq k}} Q_{kj}(t)\) represents the rates of transition from state \(k\) to all the other states \(j\) (excluding self-transitions).

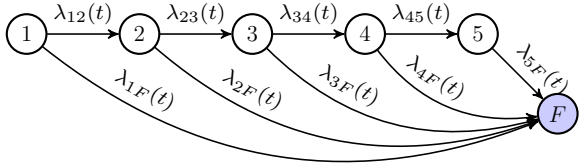
\begin{figure}[ht]
\vspace{-10pt}
\centering
\begin{tikzpicture}[->, >=stealth', auto, semithick, node distance=2.5cm, font=\sffamily\footnotesize]
\tikzset{every state/.style={fill=white, draw=black, thick, text=black, scale=0.6, font = {\Large}}}
\node[state]    (A)                    {$1$};
\node[state]    (B)[right of=A]        {$2$};
\node[state]    (C)[right of=B]        {$3$};
\node[state]    (D)[right of=C]        {$4$};
\node[state]    (E)[right of=D]        {$5$};
\node[state, fill=blue!20] (F)[below right of=E, node distance=2.5cm] {$F$};
\path
(A) edge[above]               node[pos=0.5] {$\lambda_{12}(t)$} (B)
    edge[bend right=30]        node[pos=0.2, sloped,above] {$\lambda_{1F}(t)$} (F)
(B) edge[above]               node[pos=0.5] {$\lambda_{23}(t)$} (C)
    edge[bend right=30]        node[pos=0.25, sloped, above] {$\lambda_{2F}(t)$} (F)
(C) edge[above]               node[pos=0.5] {$\lambda_{34}(t)$} (D)
    edge[bend right=30]        node[pos=0.3, sloped, above] {$\lambda_{3F}(t)$} (F)
(D) edge[above]               node[pos=0.5] {$\lambda_{45}(t)$} (E)
    edge[bend right=30]        node[pos=0.35, sloped, above] {$\lambda_{4F}(t)$} (F)
(E) edge[right=40]        node[pos=0.7, sloped, above] {$\lambda_{5F}(t)$} (F);
\end{tikzpicture}
\vspace{-15pt}
\caption{Markov chain structure for \acrshort{IHTMC}.}\label{fig:IHTMC}
\vspace{-10pt}
\end{figure}

\vspace{-15pt}
\smallpar{Pipe-element degradation model.} We define a pipe element by $K$ sequentially arranged states $S=[S_1,S_2,...,S_k]$, where $S_1$ signifies the \emph{pristine} condition and $S_k$ represents the \emph{worst condition}. This categorization is based on sewer network inspection data, which documents types of damage and their severities on a scale from 1 to 5, along with occasional instances of functional failures ($K=6$). The transitions within our \acrshort{IHTMC}, illustrated in Figure \ref{fig:IHTMC}, permit only progression from a better to a worse state, prohibiting direct improvements without repairs, while allowing any severity level to escalate to functional failure.

\vspace{-15pt}
\smallpar{Parametrization of \acrshort{IHTMC}.} We employed a parameterized approach for \acrshort{IHTMC}, involving an assumption on the hazard function. In Section \ref{sec:msdm_parametrization}, we detail the parametrization used in our experimental setup. Several aspects related to the multi-state degradation model, including hyper-parameter tuning and interval-censoring, are beyond the scope of this paper. For further information, we recommend referring to \citep{jimenez2024comparison}.

\subsection{Markov Decision Process}\label{sec:contextual_mpd}

A \acrfull{MDP} models a stochastic sequential decision process, where both costs and transition functions are dependent solely on the current state and action \citep{puterman1990markov}. Formally, an \acrshort{MDP} is described by the tuple \( \langle \mathcal{S}, \mathcal{A}, P(s_{t+1} | s_t,a_t), \mathcal{R}(s_t, a_t, s_{t+1}), \pi_0, \gamma \rangle \), with \(\mathcal{S}\) as \emph{state space}, \(\mathcal{A}\) as the \emph{action space}, \(P(s_{t+1} | s_t,a_t)\) as the \emph{transition probability function} indicating the probability of transitioning from state \(s_t\) to \(s_{t+1}\) given action \(a_t\), where \(s_t, s_{t+1} \in \mathcal{S}\) and \(a_t \in \mathcal{A}\). The \emph{reward function} \(\mathcal{R}(s_t, a_t, s_{t+1})\) specifies the reward for moving from \(s_t\) to \(s_{t+1}\) by action \(a_t\). The \emph{initial state} \(\pi_0\) represents the distribution across \(\mathcal{S}\), and \(\gamma \in [0,1]\) is the \emph{discount factor} that balances immediate versus future rewards.

\subsection{Deep Reinforcement Learning}\label{sec:rl}

Deep Reinforcement Learning (DRL) produces virtual agents that interact with environments to learn optimal behaviors through trial and error, as indicated by a reward signal \citep{arulkumaran2017deep}. DRL has found applications in robotics, video games, and navigation systems.

We utilize \acrshort{DRL} to train agents in virtual environments exhibiting degradation following the MSDM pattern, as detailed in Section \ref{sec:model}. Specifically, we apply \gls{PPO} \citep{schulman2017proximal}, a policy gradient method in \acrshort{RL}.

\acrshort{PPO} aims to optimize the policy an agent uses for action selection, maximizing expected returns. It addresses stability and efficiency issues encountered in previous algorithms like \textit{Trust Region Policy Optimization} by offering a simpler and less computationally expensive method to ensure minor policy updates.

This is achieved through an innovative objective function that penalizes significant deviations from the previous policy, fostering stable and consistent learning. The term ``proximal'' denotes maintaining proximity between the new and old policies, facilitating a stable training process and rendering PPO popular across various \acrshort{RL} applications.
\section{Methodology}\label{sec:methodology}

\begin{figure*}[ht]
 \centering
        \includegraphics[width=\linewidth]{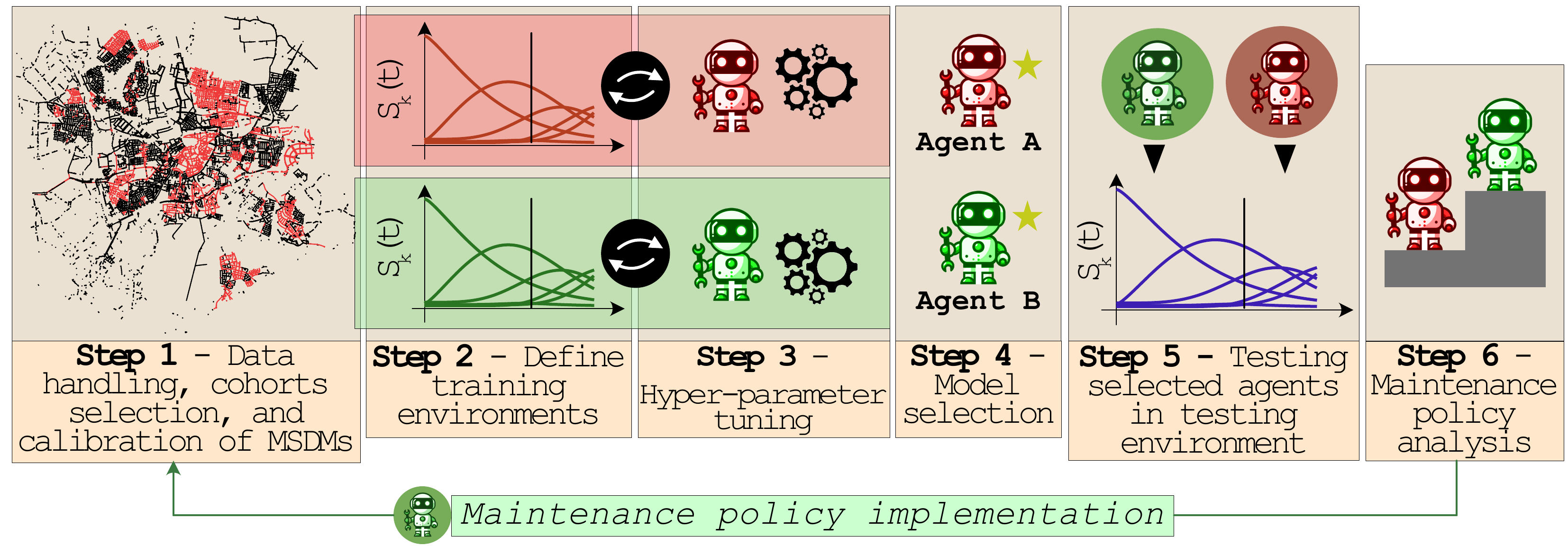}
        \caption{Methodology overview for sewer pipe maintenance policy optimization using Deep Reinforcement Learning and Multi-State Degradation models.}
        \label{fig:methodology_v2}
\end{figure*}

Our methodology, illustrated in Figure \ref{fig:methodology_v2}, comprises six steps, detailed below.

\setlength{\labelsep}{5pt} 
\begin{compactenum}[\bfseries Step 1.]
\item Perform data handling of historical inspection records, selecting subsets (cohorts) of interest, and calibrating the MSDM on this data. This step is beyond the scope of this paper; for details, see \cite{jimenez2022deterioration,jimenez2024comparison}. The results of this step are given in Section~\ref{sec:msdm}.

\item After calibrating the MSDM, integrate these models into an environment suitable for RL applications. We present the details of our environment integrating MSDM in Section~\ref{sec:model}. In addition, we define environments for training RL agents. This is to test different MSDM hypotheses; details on this can be found in Section~\ref{sec:experimental_setup}.

\item Train \acrshort{DRL} agents with \acrshort{PPO}. Use \texttt{optuna} for hyper-parameter tuning and \texttt{Stable Baselines3} for RL implementation. Details are in Section~\ref{sec:hyper_param_tunning}.

\item Train and select the RL agents with the optimal hyper-parameters on the \textit{training} environments. In essence, these agents learn the dynamics described by the MSDM encoded in the environment.

\item \chl{Compare the maintenance policies advised by the} \acrshort{RL} \chl{agents using the \textit{test} environment against the heuristics}: Condition-Based Maintenance (CBM), Scheduled Maintenance (SchM), and Reactive Maintenance (RM). Find the definition of these heuristics in Section~\ref{sec:comparison_heuristics}.
 
\item Analyze and compare the behavior of the maintenance strategies for the different RL models and heuristics. Reflect on the policies advantages and disadvantages. Find in Section~\ref{sec:policy_analysis_overview} the overview of this comparison, and in Section~\ref{sec:policy_analysis_over_episode} are the details along episodes.
\end{compactenum}

\section{Multi-state degradation models}\label{sec:msdm}

\subsection{Case study}\label{sec:case_study}
Our case study conducts a detailed examination of the sewer pipe network in Breda, the Netherlands, which comprises 25,727 sewer pipes covering 1,052 km, mostly built after 1950. The network is primarily made of concrete (72\%) and PVC (27\%), with the shapes of the pipes being predominantly round (95\%) and ovoid (5.4\%). These pipes are designed for transportation (98.2\%), with 88\% being up to 60 meters in length. Additionally, 98.3\% have a diameter of up to 1 meter, with the most common diameter being 0.2 meters, and they carry mixed (63\%), rain (21\%), and waste (16\%) contents. The condition of the pipes is evaluated through visual inspections according to the European standard EN 13508 \citep{EN13508Part1, EN13508Part2}, focusing on identifying and classifying damage with specific codes. This study specifically addresses the damage code \textit{BAF}, which signifies \textit{surface damage} and was \chl{observed} in 35.3\% of the inspections.

\subsection{Parametrization}\label{sec:msdm_parametrization}

We consider three distributions for hazard rate functions: Exponential, Gompertz, and Weibull. The hazard rates \( \lambda(t|\cdot) \) for these distributions are specified as follows:

\vspace{-10pt}
\begin{subequations}\label{eq:rate_functions}
\begin{align}
    \text{\emph{Exponential} function:} \quad & \lambda^E(t|\epsilon) = \epsilon, \label{eq:exponential_rate} \\
    \text{\emph{Gompertz} function:} \quad & \lambda^G(t| \alpha, \beta) = \alpha \beta e^{\beta t} \label{eq:gompertz_rate} \\
    \text{\emph{Weibull} function:} \quad & \lambda^W(t| \eta,\rho) = \frac{\rho}{\eta} \Big( \frac{t}{\eta} \Big)^{\rho-1}\label{eq:weibull_rate}
\end{align}
\end{subequations}

In Eq.~(\ref{eq:exponential_rate}), a constant hazard rate indicates that the degradation model assumes a \emph{homogeneous} time, exhibiting \emph{memoryless} properties. Eq.~(\ref{eq:gompertz_rate}) and Eq.~(\ref{eq:weibull_rate}) present varying hazard rates, which indicates \emph{inhomogeneous} time.

\subsection{Solving the Multi-State Degradation Model} 
In Figure~\ref{fig:IHTMC}, we defined the structure of the Markov chain to model degradation in a sewer pipe, and in Section \ref{sec:msdm_parametrization} we introduced the hazard rate functions. In the following, we present the corresponding system of differential equations.
\begin{subequations}
\begin{align}
\frac{\partial S_1(t)}{dt} &= - \big(\lambda_{12}(t|\cdot) + \lambda_{1F}(t|\cdot)\big)S_1(t) \\
\frac{\partial S_2(t)}{dt} &= \lambda_{12}(t|\cdot)S_1(t) - \big(\lambda_{23}(t|\cdot) + \lambda_{2F}(t|\cdot)\big)S_2(t) \\
\frac{\partial S_3(t)}{dt} &= \lambda_{23}(t|\cdot)S_2(t) - \big(\lambda_{34}(t|\cdot) + \lambda_{3F}(t|\cdot)\big)S_3(t) \\
\frac{\partial S_4(t)}{dt} &= \lambda_{34}(t|\cdot)S_3(t) + \big(-\lambda_{45}(t|\cdot) - \lambda_{4F}(t|\cdot)\big)S_4(t) \\
\frac{\partial S_5(t)}{dt} &= \lambda_{45}(t|\cdot)S_4(t) - \lambda_{5F}(t|\cdot)S_5(t) \\
\frac{\partial S_F(t)}{dt} &= \lambda_{1F}(t|\cdot)S_1(t) + \lambda_{2F}(t|\cdot)S_2(t) + \lambda_{3F}(t|\cdot)S_3(t) \nonumber \\
&\quad + \lambda_{4F}(t|\cdot)S_4(t) + \lambda_{5F}(t|\cdot)S_5(t)
\end{align}
\label{eq:sys_diff_eqs}
\end{subequations}

Eq. \ref{eq:sys_diff_eqs} is solved using  numerical methods, specifically the \texttt{LSODA} algorithm from the FORTRAN \texttt{odepack} library implemented in SciPy \citep{Eric2001}. This algorithm solves systems of ordinary differential equations by employing the \texttt{Adams/BDF} method with automatic stiffness detection.

\subsection{Parametric \texorpdfstring{\acrlongpl{MSDM}}{MSDMs}}

We extract a subset from our case study data set to construct a cohort with concrete sewer pipes carrying \textit{mixed and waste content} (cohort \texttt{CMW}), representing 37.1\% of the sewer network. The model parameters for this cohort are detailed in \ref{app:msdm_parameters} in Tables \ref{tab:hyperparameter_msdm_CMW} and \ref{tab:initial_state_vector_CWM}.

\vspace{-5pt}
\begin{figure}[!h]
    \centering
    \includegraphics[width=\linewidth]{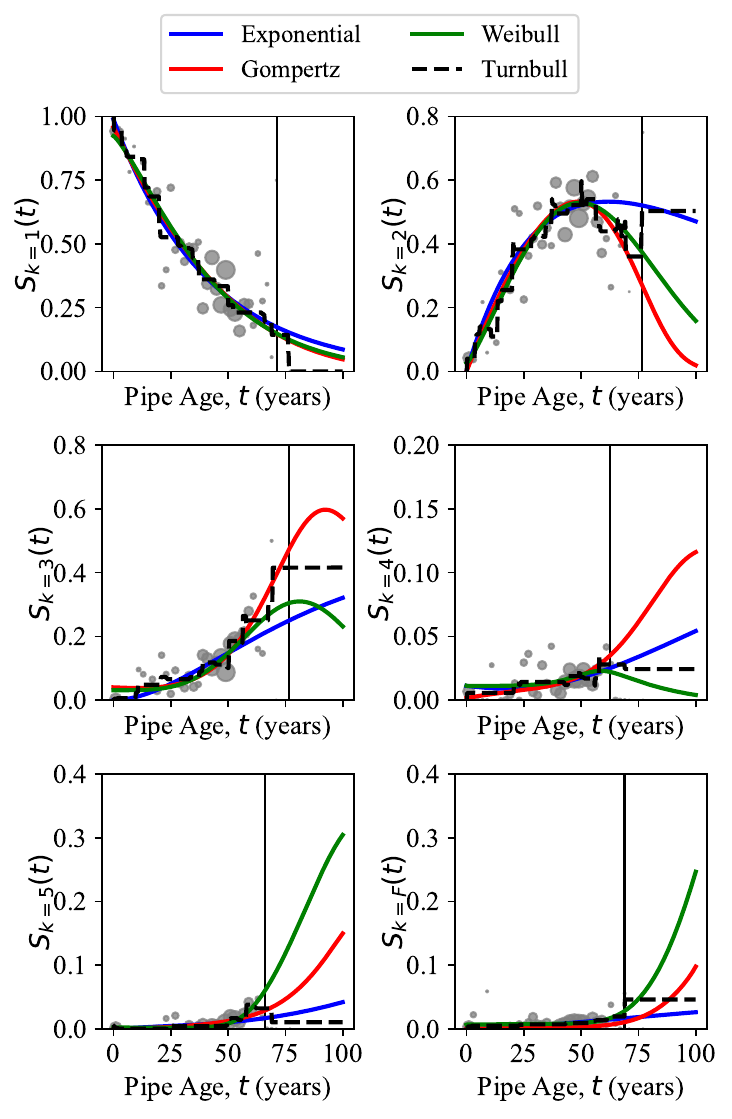}
    \vspace{-15pt}
    \caption{Probability of being in state \(k \in \Omega\) at pipe age \(t\) \(S_k(t)\), using three hazard functions modeled via Exponential, Gompertz, and Weibull probability density functions. The Turnbull non-parametric estimator indicates the ground truth. The gray circles indicate the frequency based on the inspection data set.}
    \vspace{-15pt}
    \label{fig:msdm_CMW}
\end{figure}

Figure~\ref{fig:msdm_CMW} illustrates the \acrshortpl{MSDM} predictions, detailing the stochastic dynamics of sewer pipe degradation for pipes in cohort \texttt{CMW}. As Figure~\ref{fig:IHTMC} describes, this degradation is segmented into five sequentially ordered severity levels ($k=1$ to $k=5$), plus a functional failure state ($k=F$). Differences in the y-axis scales are intentional, to emphasize details and behaviors that various degradation models express across severity levels.

Gray circles represent the frequency per severity level from the inspection dataset. \cite{jimenez2022deterioration} details how these frequencies are computed. Vertical black lines in Figure~\ref{fig:msdm_CMW} mark the last available data point for each severity level.

Additionally, Figure \ref{fig:msdm_CMW} presents the \textit{Turnbull} non-parametric estimator, which assumes no specific distribution for survival times \citep{turnbull1976empirical}. In our context, this estimator represents the ground truth of stochastic degradation behavior in sewer pipes.

Tables \ref{tab:error_CMW} presents the Root Mean Square Error (RMSE) computed with respect to the Turnbull estimator, for each \acrshort{MSDM} assumption, for cohorts \texttt{CMW}. These results show that models employing Gompertz and Weibull distributions yield smaller RMSEs compared to the one using the Exponential distribution.

\begin{table}[!h]
\centering
\caption{RMSE with respect Turnbull estimator, per severity level $k$ and total RMSE, cohort:~\texttt{CMW}.}
\begin{tabular}{rrrr}
\toprule
 &  Exponential &  Gompertz &  Weibull \\
\midrule
$S_{k=1}(t)$    &     3.38E-02 &  3.27E-02 & 3.34E-02 \\
$S_{k=2}(t)$    &     7.04E-02 &  3.70E-02 & 3.57E-02 \\
$S_{k=3}(t)$    &     6.27E-02 &  2.81E-02 & 4.38E-02 \\
$S_{k=4}(t)$    &     4.28E-03 &  1.13E-02 & 5.06E-03 \\
$S_{k=5}(t)$    &     8.33E-03 &  1.09E-02 & 3.04E-02 \\
$S_{k=F}(t)$    &     9.19E-03 &  1.17E-02 & 3.62E-03 \\
\addlinespace
Total &     4.13E-02 &  2.45E-02 & 2.96E-02 \\
\bottomrule
\end{tabular}
\label{tab:error_CMW}
\end{table}

These \acrshortpl{MSDM} serve two crucial roles within our environment: first, they drive the degradation behavior of sewer pipes, effectively emulating how sewer pipes degrade over time. Second, the output from the \acrshortpl{MSDM} is incorporated as prognostic information, available to the agent to support decisions at any time point. This latter aspect is considered a novel feature of our framework. Details on the \acrshort{MDP} are provided in the section below.

\section{Definition of Markov Decision Process for Maintenance Policy Optimization of a Sewer Pipe considering pipe length degradation}\label{sec:model}

\begin{figure*}[ht]
 \centering
        \includegraphics[width=0.775\linewidth]{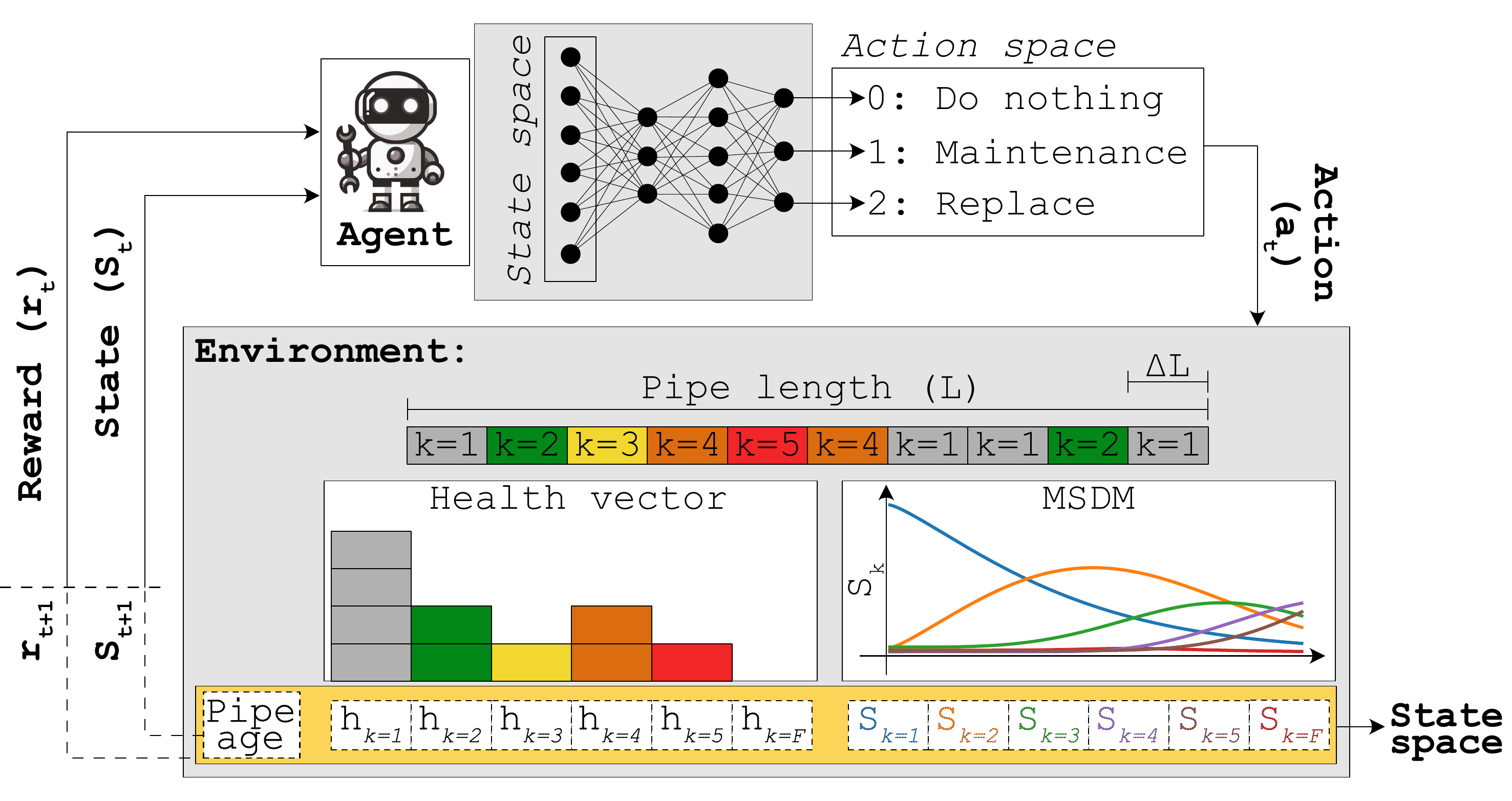}
        \caption{Environment for maintenance policy optimization of a sewer pipe via Deep Reinforcement Learning, considering degradation along the pipe length.}
        \label{fig:RL_framework}
\end{figure*}

Figure \ref{fig:RL_framework} provides the workflow that the RL agent uses to learn maintenance policies for sewer pipes, considering degradation along the pipe length. 
In the following sections, we provide the details of the environment, namely the state and action spaces, as well as the transition probability and reward functions.

\subsection{State space (\( \mathcal{S} \))}\label{sec:state_space}

Our approach focuses on developing age-based maintenance policies, incorporating the sewer pipe's age into the state representation. Our state space is \textit{continuous} and it is structured to include three key components: (i) the age of the pipe, (ii) the \textit{health vector}, and (iii) the stochastic prediction of severity levels. We next describe the last two components.

\subsubsection{Health vector (\textbf{h})}

In modeling the degradation of linear structures like sewer pipes, it is essential to represent changes accurately along their length. For this purpose, we define a \textit{health vector} (\(\textbf{h}\)), which quantitatively measures the degradation at various points along the pipe. The vector is crucial in our framework, particularly influencing the reward function as described in Section \ref{sec:reward_function}.

\vspace{-20pt}
\paragraph{Construction of \(\textbf{h}\):} 
We discretize the pipe into segments of equal length \( \Delta L \), with \( \Delta L < L \), where \( L \) is the total length of the pipe. The number of segments, \( n_d \), is calculated using the ceiling function to ensure it remains an integer even if \( L \) is not perfectly divisible by \( \Delta L \):
\vspace{-5pt}
\begin{equation}
n_d = \left\lceil \frac{L}{\Delta L} \right\rceil
\end{equation}
\vspace{-10pt}

Each segment's degradation level is initially assessed and categorized into \textit{severity levels} according to the \acrshort{MSDM}. As the degradation progresses, the state of each segment changes following the transition probabilities described by the matrix \( P_{i,j} \), where \( i \) is the current severity level, and \( j \) is the subsequent severity level, as described by the forward Kolmogorov equation (Eq. \ref{eq:forward_kolmogorov}).

\chl{Notice that by doing this, we assume there is no statistical dependency between segments, which is a strong assumption that needs further research. However, for simplicity, we maintain this assumption in our degradation model.}

\vspace{-20pt}
\paragraph{Quantifying Degradation:}
The distribution of severity levels across the pipe is captured in vector \(\textbf{d}\), with each element indicating the severity level of a segment. To quantify this distribution in the health vector \(\textbf{h}\), we first count the number of segments at each severity level \( k \) using the following expression:
\vspace{-5pt}
\begin{equation}
n_{d_k} = \sum_{i=1}^{n_d} \mathbf{1}_{\{\textbf{d}_i = k\}}
\end{equation}
where \(\mathbf{1}\) is the indicator function that is 1 if the condition is true and 0 otherwise. The health vector \(\textbf{h}\) is then determined by normalizing these counts to reflect the proportion of segments at each severity level:
\vspace{-5pt}
\begin{equation}
\textbf{h}_k = \frac{n_{d_k}}{n_d}
\label{eq:damage_points}
\end{equation}
\vspace{-10pt}

Here, \( n_{d_k} \) is the number of segments at severity level \( k \). Thus, \(\textbf{h}_k\) becomes part of the state space indicating the \textit{level of degradation} present in the pipe.

\subsubsection{Stochastic prediction of severity levels}

To enable the agent to access information provided by the \acrshort{MSDM}, we incorporate the prediction of severity levels into the state space. This is accomplished by solving Eq. \ref{eq:master_MC_equation}, yielding a distribution \(S_{k}(t)\).

Finally, our state space is defined as a tuple with 13 elements:
\vspace{-3pt}
\begin{equation*}
\mathcal{S}  =   \langle \textcolor{darkgreen}{\text{Pipe Age}}, \textcolor{red}{\textbf{h}_{1}}, \textcolor{red}{\textbf{h}_{2}}, \textcolor{red}{\textbf{h}_{3}}, \textcolor{red}{\textbf{h}_{4}}, \textcolor{red}{\textbf{h}_{5}}, \textcolor{red}{\textbf{h}_{F}}, \textcolor{blue}{S_{1}}, \textcolor{blue}{S_{2}}, \textcolor{blue}{S_{3}}, \textcolor{blue}{S_{4}}, \textcolor{blue}{S_{5}}, \textcolor{blue}{S_{F}} \rangle 
\end{equation*}
\subsection{Action space (\( \mathcal{A} \))}\label{sec:action_space}

Our action space \( \mathcal{A} \) is \textit{discrete} with dimensionality \( | \mathcal{A} | = 3 \). At each time step \(t\), the agent selects an action \( a_t \). If the decision at time \(t\) is \textit{do nothing}, \(a_t\) is set to \(0\). To perform \textit{maintenance}, \(a_t\) is set to \(1\), and to \textit{replace} the pipe, \(a_t\) is set to \(2\). The outcomes of these actions are discussed in Section~\ref{sec:transition_function}.
\subsection{Transition function  (\( P \))}\label{sec:transition_function}

Our transition function \(P(s_{t+1} | s_t , a_t)\) is \textit{stochastic}, dependent on time \(t\), and considers both the actions \( a \in \mathcal{A} \) and the current \(s_t\) and next state \(s_{t+1}\) dynamics described by the \acrshort{MSDM}. We illustrate the behavior of \(P\) with the following example.

For a 30-year-old pipe with length \(L = 40\) meters and discretized in segments of length \(\Delta L = 1\), let the current state space be \(s_{t=30} \in \mathcal{S}\):
\begin{equation*}
\begin{aligned}
s_{t=30}  =   \left\langle  \textcolor{darkgreen}{30}, \textcolor{red}{0.60}, \textcolor{red}{0.35}, \textcolor{red}{0.025}, \textcolor{red}{0.025}, \textcolor{red}{0.0}, \textcolor{red}{0.0}, \right. \\
\left. \textcolor{blue}{0.475}, \textcolor{blue}{0.436}, \textcolor{blue}{0.069}, \textcolor{blue}{0.010}, \textcolor{blue}{0.005}, \textcolor{blue}{0.005} \right\rangle.
\end{aligned}
\end{equation*}
\( s_{t=30} \) indicates the age of the pipe is \textcolor{darkgreen}{30} years. From Eq. \ref{eq:damage_points}, the number of segments at severity $k$ is determined by multiplying the health vector (\textcolor{red}{\(\textbf{h}_k\)}):
\begin{equation*}
\textcolor{red}{\textbf{h}_k} = [\textcolor{red}{0.60}, \textcolor{red}{0.35},\textcolor{red}{0.025}, \textcolor{red}{0.025}, \textcolor{red}{0.0}, \textcolor{red}{0.0}]
\end{equation*}
by 40 meters, yielding \( \textcolor{red}{n_{d_k}} = [24, 14, 1, 1, 0, 0] \), indicating that, out of the 40 meters of pipe length, 24 segments of 1 meter are at severity \(k=1\), 14 at severity \(k=2\), and so forth.

The distribution \textcolor{blue}{\( S_k(t=30.0) \)} predicts the probability of being in a severity level $k$ at age \(t=30\). This is achieved by evaluating \(t=30.0\) in the corresponding \acrshort{MSDM}.
\begin{equation*}
 \textcolor{blue}{S_k(t=30.0)} = [\textcolor{blue}{0.475}, \textcolor{blue}{0.436}, \textcolor{blue}{0.069}, \textcolor{blue}{0.010}, \textcolor{blue}{0.005}, \textcolor{blue}{0.005}]
\end{equation*}
Assuming the agent takes an action every half year, we illustrate the effect of each action in \( \mathcal{A} \) below.

\begin{compactenum}[-]
    \item If \textbf{\( a_t=0 \)}: the agent decides to ``do nothing'', the pipe's degradation evolves in line with the \acrshort{MSDM} progression. Here the new state space becomes \( s^{a = 0}_{t=30.5}  \).
    \begin{equation*}
    \begin{aligned}
    s^{a = 0}_{t=30.5}  =   \langle & \textcolor{darkgreen}{30.5}, \textcolor{red}{0.575}, \textcolor{red}{0.35}, \textcolor{red}{0.05}, \textcolor{red}{0.025}, \textcolor{red}{0.0}, \textcolor{red}{0.0}, \\
    &  \textcolor{blue}{0.470}, \textcolor{blue}{0.439}, \textcolor{blue}{0.071}, 
    \textcolor{blue}{0.010}, \textcolor{blue}{0.05}, \textcolor{blue}{0.05} \rangle 
    \end{aligned}
    \end{equation*}
    Notice that the pipe age increased to \textcolor{darkgreen}{30.5}, and \( \textcolor{red}{n_{d_k}} = [23, 14, 2, 1, 0, 0] \), where a segment with severity \(k=1\) progressed to \(k=2\), and one segment with \(k=2\) advanced to \(k=3\). Additionally, \textcolor{blue}{\( S_k(t) \)} is updated by evaluating \(t=30.5\).

    \item If \( a_t = 1 \): the agent decides to ``perform maintenance,'' all damage points with severity levels \( k \in \{3, 4, 5\} \) are moved to \(k=2\). Consequently, this action does not affect damage points with severity levels \( k \in \{1, 2, F\} \).  The new state space becomes \( s^{a = 1}_{t=30.5}  \).
    \begin{equation*}
    \begin{aligned}
    s^{a = 1}_{t=30.5}  =   \langle & \textcolor{darkgreen}{30.5}, \textcolor{red}{0.60}, \textcolor{red}{0.40}, \textcolor{red}{0.0}, \textcolor{red}{0.0}, \textcolor{red}{0.0}, \textcolor{red}{0.0}, \\
     &  \textcolor{blue}{0.47}, \textcolor{blue}{0.439}, \textcolor{blue}{0.071}, 
    \textcolor{blue}{0.010}, \textcolor{blue}{0.05}, \textcolor{blue}{0.05} \rangle 
    \end{aligned}
    \end{equation*}
    Notice that the pipe age increased to \textcolor{darkgreen}{30.5}, and \( \textcolor{red}{n_{d_k}} = [24, 16, 0, 0, 0, 0] \). However, \( \textcolor{blue}{S_k(t)} \) is updated by evaluating \(t=30.5\), same as when \( a_t = 0 \).

    \item If \(a_t = 2\): the agent decides to ``replace'' the pipe, resetting its condition to as good-as-new. The new state space is \(s^{a = 2}_{t=0.0}\):
    \begin{equation*}
    \begin{aligned}
    s^{a = 2}_{t=0.0} = \langle & \textcolor{darkgreen}{0.0}, \textcolor{red}{1.0}, \textcolor{red}{0.0}, \textcolor{red}{0.0}, \textcolor{red}{0.0}, \textcolor{red}{0.0}, \textcolor{red}{0.0}, \\
    & \textcolor{blue}{0.986}, \textcolor{blue}{0.014}, \textcolor{blue}{0.0}, \textcolor{blue}{0.0}, \textcolor{blue}{0.0}, \textcolor{blue}{0.0} \rangle.
    \end{aligned}
    \end{equation*}
    The pipe age is reset to \textcolor{darkgreen}{0.0}, with \textcolor{red}{\(n_{d_k}\)} \(= [40, 0, 0, 0, 0, 0]\), and \textcolor{blue}{\(S_k(t)\)} is updated for \(t=0.0\).
\end{compactenum}
\subsection{Reward function (\( \mathcal{R} \))}\label{sec:reward_function}

Our reward function \( \mathcal{R}(s_t, a_t, s_{t+1}) \) assigns a reward \( r_t\) at every decision point \(t\), determined by the current state \( s_t \) and action \( a_t \). This function integrates the costs of maintenance (\( C_M \)), replacement (\( C_R \)), and failures~(\( C_F \)). \(\mathcal{R}\) \chl{is \textit{sparse} because it issues a non-zero value only when failures occur or interventions are undertaken.}

Maintenance cost \( C_M \) is calculated as per Eq. \ref{eq:maintenance_costs}, where it combines a variable cost based on severity \(k\) with a fixed logistic cost of €500, covering the expenses related to maintenance. 

These costs vary with the severity level $k$, as detailed in Table~\ref{tab:maintenance_costs}. Note that no maintenance costs are associated with \(k=F\) because maintenance cannot be performed on a segment that has already failed. In this case, the agent must replace.

\begin{equation}
C_M  = - (\textbf{h}_k \cdot c_M^{k} + 500)
\label{eq:maintenance_costs}
\end{equation}
\vspace{-10pt}
\begin{table}[!ht]
\small
\centering
\caption{Maintenance costs per severity $k$ per segment~(\(c_M^{k}\))}
\label{tab:maintenance_costs}
\begin{tabular}{ccccccc}
\toprule
& $k=1$ & $k=2$ & $k=3$ & $k=4$ & $k=5$  & $k=F$ \\
\midrule
 \(c_M^{k}=\)  & 0 & 0 & -\texteuro{}500 & -\texteuro{}700 & -\texteuro{}900  & N.A.\\
\bottomrule
\end{tabular}
\vspace{-10pt}
\end{table}

Replacement costs (\( C_R \)) is computed with Eq. \ref{eq:replacement_cost}:
\begin{equation}
C_R  =  -(450 + 0.66D + 0.0008D^2)L
\label{eq:replacement_cost}
\end{equation}
Here, $L$ and $D$ denote the pipe's length in meters and diameter in millimetres, respectively. \( C_R  \) is in Euros (\texteuro{}). 

The cost of failure, denoted by \(C_F\), entails assigning a substantial penalty when the agent allows a segment of the pipe to achieve a failure state (\(k=F\)). This penalty cost is established at \texteuro{}-100,000. Our reward function is then:
\begin{equation}
r_t = \frac{C_M + C_R + C_F}{100'000+900\times40}  = \frac{C_M + C_R + C_F}{136'000}
\end{equation}
where \(r_t\) represents the reward obtained at time \(t\), the normalization constant \(136'000\) corresponds to the most expensive penalty possible at time \(t\). Thus, \(r_t\) is defined within the interval \([-1, 0]\). This reward function aims for the agent to balance maintenance actions with the prevention of undesirable pipe conditions.

\section{Experimental setup}\label{sec:experimental_setup}

\subsection{Setup}

We will evaluate our framework with a single pipe of constant length (40 meters) and diameter (200 mm) from the cohort \texttt{CMW}, which carries mixed and waste content. Given the constant dimensions, the replacement cost \(C_R\), as defined in Eq.~\ref{eq:replacement_cost}, is \texteuro{}24,560. \chl{The pipe age, when initializing the episode, is randomly sampled from the uniform distribution} \(U \sim [0, 50]\)\chl{, allowing the agent to learn the behavior of pipes within this age range.
Additionally, we evaluate the policy in steps of half a year and} \(\Delta L = 1 \text{ meter} \).

In the methodology section, we describe the training of two agents: \textbf{Agent-E} and \textbf{Agent-G}. \textbf{Agent-E} is trained in an environment where sewer pipe degradation follows the \acrshort{MSDM} parameterised with an \textit{Exponential} probability density function, while \textbf{Agent-G} is trained in an environment where degradation follows the \acrshort{MSDM} parameterised with a \textit{Gompertz} probability density function.

Both agents are tested in an environment where sewer pipe degradation follows the \acrshort{MSDM} parameterized with the \textit{Weibull} probability density function.

During training, each agent follows a specific state space, defined as follows:

\vspace{-5pt}
\begin{subequations}
\begin{equation}
\mathcal{S^{\textbf{Agent-E}}_\text{Training}} = \langle \textcolor{darkgreen}{\text{Pipe Age}}, \textcolor{red}{\textbf{h}_{k}^{E}}, \textcolor{blue}{S_{k}^{E}(t)} \rangle
\end{equation}
\begin{equation}
\mathcal{S^{\textbf{Agent-G}}_\text{Training}} = \langle \textcolor{darkgreen}{\text{Pipe Age}}, \textcolor{red}{\textbf{h}_{k}^{G}}, \textcolor{blue}{S_{k}^{G}(t)} \rangle
\end{equation}
\label{eq:training_setup}
\end{subequations}

Here, \(\mathcal{S}\) represents the state space for each agent during training. The subscripts \(E\) and \(G\) denote the \textit{Exponential} and \textit{Gompertz} probability density functions, respectively. Each agent's objective is to learn an optimal maintenance strategy based on their environment's dynamics.

For testing, both agents are evaluated in the same environment, with the state space defined as follows:

\vspace{-10pt}
\begin{subequations}
\begin{equation}
\mathcal{S^{\textbf{Agent-E}}_\text{Testing}} = \langle \textcolor{darkgreen}{\text{Pipe Age}}, \textcolor{red}{\textbf{h}_{k}^{W}}, \textcolor{blue}{S_{k}^{E}(t)} \rangle
\end{equation}
\begin{equation}
\mathcal{S^{\textbf{Agent-G}}_\text{Testing}} = \langle \textcolor{darkgreen}{\text{Pipe Age}}, \textcolor{red}{\textbf{h}_{k}^{W}}, \textcolor{blue}{S_{k}^{G}(t)} \rangle
\end{equation}
\label{eq:testing_setup}
\end{subequations}

In both cases, \(S_{k}^{E}(t)\) and \(S_{k}^{G}(t)\) remain consistent with the training phase, reflecting the \acrshort{MSDM} predictions. However, the health vector \(\textbf{h}_k\) follows the degradation behavior described by the \textit{Weibull} probability density function, indicated by the subscript \(W\).

\subsection{Comparison of maintenance strategies}\label{sec:comparison_heuristics}

We compare the RL agent's performance against maintenance policies based on heuristics. For this, we define the following:

\begin{compactitem}
\item \textbf{Condition-Based Maintenance (CBM)}: Maintenance actions are based on the sewer pipe's condition. Specifically, replacement (\(a_t=2\)) is performed if \( \texttt{pipe\_age} \geq 70 \) or \( \text{\textbf{h}}_{k=F} \geq 0.0 \); maintenance (\(a_t=1\)) is conducted if \( \text{\textbf{h}}_{k=4} \geq 0.1 \) or \( \text{\textbf{h}}_{k=5} \geq 0.05 \); otherwise, no action (\(a_t=0\)) is taken.

\item \textbf{Scheduled Maintenance (SchM)}: Actions are time-based. Replacement (\(a_t=2\)) is executed if \( \text{\textbf{h}}_{k=F} \geq 0.0 \); maintenance (\(a_t=1\)) occurs every 10 years; otherwise, no action (\(a_t=0\)) is taken.

\item \textbf{Reactive Maintenance (RM)}: Replacement is undertaken only upon pipe failure, i.e., replacement (\(a_t=2\)) is performed if \( \text{\textbf{h}}_{k=F} \geq 0.0 \); otherwise, no action (\(a_t=0\)) is taken.
\end{compactitem}

Note that CBM and SchM are defined based on plausible values. However, these heuristics can be further calibrated for enhanced performance, which is beyond the scope of this paper.
\section{Results}\label{sec:results}
\subsection{Implementation and hyper-parameter tuning}\label{sec:hyper_param_tunning}

Our framework uses \texttt{Stable Baselines3}~\citep{raffin2021stable}, comprising robust implementations of RL algorithms in \chl{PyTorch}~\citep{ansel2024pytorch}. Specifically, we utilize the \acrshort{PPO} algorithm. Hyper-parameter optimization is performed using \texttt{optuna} \citep{optuna_2019}, a framework dedicated to automating the optimization of hyper-parameters.

The search space encompasses: \chl{exponentially-decaying learning rate with a decay rate of 0.05, with an initial learning rate ranging from} \(10^{-5}\) to \(10^{-2}\), discount factor (\(\gamma\)) from 0.8 to 0.9999, entropy coefficient from 0.0001 to 0.01, steps per update (\texttt{n\_steps}) from 250 to 3000, batch sizes from 16 to 256, activation functions (`tanh', `relu', `sigmoid'), policy network architectures ([16, 16], [32, 32], [64, 64], [32, 32, 32]), and training epochs (\texttt{n\_epochs}) from 5 to 100.

We set up \texttt{optuna} to conduct 500 trials, aiming to maximise cumulative reward in 100 episodes. Table \ref{tab:hyperparameters} details the optimal hyper-parameters identified. These parameters are used to obtain the results discussed in Sections \ref{sec:policy_analysis_overview} and \ref{sec:policy_analysis_over_episode}, where our agents are trained over a total of 5 million time steps.

\begin{table}[ht]
\centering
\caption{Optimal hyper-parameters found using \texttt{optuna}.}
\begin{tabular}{rc}
\hline
Hyper-parameter & Value \\
\hline
Learning rate & 0.0003 \\
Discount factor & 0.995 \\
Entropy coefficient & 0.008 \\
Steps per update (\texttt{n\_steps}) & 2080 \\
Batch size & 104 \\
Activation function & Sigmoid \\
Policy network architecture & [32, 32, 32] \\
Training epochs (\texttt{n\_epochs}) & 50 \\
\hline
\end{tabular}
\label{tab:hyperparameters}
\end{table}
\subsection{Policy analysis: overview}\label{sec:policy_analysis_overview}

This section offers a broad evaluation of the policies, with a detailed analysis over episodes presented in Section \ref{sec:policy_analysis_over_episode}. We compare the agents' performances with the heuristics detailed in Section \ref{sec:comparison_heuristics} across 100 simulations in the \textbf{test} environment (Eq. \ref{eq:testing_setup}), considering pipe ages of 0, 25, and 50 years, aiming to evaluate policy efficacy concerning degradation over varying pipe ages.

Table \ref{tab:tab_comparison_mean_policy_cost} presents the \textit{mean policy cost} for Agent-E, Agent-G, CBM, SchM, and RM, highlighting the best and second-best policies in \textcolor{blue}{blue} and \textcolor{red}{red}, with corresponding means and standard deviations from the simulations.

\begin{table}[ht]
\centering
\small
\caption{Policy cost comparison: Mean and standard deviation (Std.) of costs for Agent-E, Agent-G, CBM, SchM, and RM, evaluated over 100 episodes in the test environment. Costs, in thousands of Euros (\texteuro{}), for pipe ages of 0, 25, and 50 years.}
\begin{tabular}{rcccccc}
\toprule
& \multicolumn{2}{c}{Pipe age: 0}
& \multicolumn{2}{c}{Pipe age: 25} & \multicolumn{2}{c}{Pipe age: 50} \\
\cmidrule(r){2-3} \cmidrule(lr){4-5} \cmidrule(l){6-7}
Policy &  Mean &    Std. &   Mean &    Std. &   Mean &    Std. \\
\midrule
Agent-E &  51.3 &   80.8 &  116.5 &  97.7 &  156.8 &  121.2 \\
Agent-G &  \textcolor{blue}{\textbf{39.7}} &   66.2 &   \textcolor{blue}{\textbf{78.7}}  &  96.6 &  \textcolor{red}{\textbf{127.1}}   &  128.3 \\
CBM     &  51.3 &  107.2 &  112.3 &  88.5 &   \textcolor{blue}{\textbf{110.7}} &   86.6 \\
SchM   &  \textcolor{red}{\textbf{42.5}}  &   70.9 &  \textcolor{red}{\textbf{78.9}}  &  96.4 &  159.8 &   95.9 \\
RM      &  48.6 &   76.6 &  135.8 &  86.5 &  165.7 &   80.8 \\
\bottomrule
\end{tabular}
\label{tab:tab_comparison_mean_policy_cost}
\end{table}

From these results, we observe that Agent-G's policy generally outperforms others for pipe ages of 0 and 25 years, securing a second-best position for pipes aged 50 years. It is noted that the cost of all policies increases with pipe age, which aligns with expectations as older pipes require more interventions.

After reviewing the mean policy costs, our focus shifts to the specific actions involved in each policy. Table~\ref{tab:mean_number_actions_per_policy} provides a summary of the actions executed by each policy across simulations for different pipe ages. For new pipes, the SchM policy leads in maintenance activities (\(a_t=1\)), with Agent-G following. In terms of replacements (\(a_t=2\)), Agent-E is the foremost in implementing this action, with CMB in second place. Both Agent-G and SchM exhibit lower replacement frequencies, explaining the mean policy costs since maintenance actions incur lower expenses compared to the penalties and replacement costs resulting from pipe failures.

{
\setlength{\tabcolsep}{4.3pt} %
\begin{table}[!h]
\centering
\small
\caption{Percentage of actions per policy obtained with Agent-E, Agent-G, CBM, SchM, and RM, evaluated over 100 episodes in the test environment, for different pipe ages.}
\vspace{-5pt}
\begin{tabular}{ccrrrrr}
\toprule
Pipe age & Action &  Agent-E &  Agent-G & CBM &  SchM & RM \\
\midrule
\multirow{3}{*}{0} & \(a_t=0\)  & 99.5 &    97.51 &  99.54 &  94.76 &  99.61 \\
                   & \(a_t=1\)  & 0.0 &     2.21 &   0.05 &   4.95 &   0.00 \\
                   & \(a_t=2\)  & 0.5 &     0.28 &   0.41 &   0.29 &   0.39 \\                   
\addlinespace
\multirow{3}{*}{25} & \(a_t=0\)  & 98.81 &    94.96 &  98.14 &  94.56 &  98.92 \\
                    & \(a_t=1\)  & 0.00 &     4.50 &   0.62 &   4.94 &   0.00 \\
                    & \(a_t=2\)  & 1.19 &     0.53 &   1.24 &   0.50 &   1.08 \\

\addlinespace
\multirow{3}{*}{50} & \(a_t=0\)  & 98.4 &    94.52 &  98.05 &  93.99 &  98.68 \\
                    & \(a_t=1\)  & 0.0 &     4.43 &   0.67 &   4.88 &   0.00 \\
                    & \(a_t=2\)  & 1.6 &     1.05 &   1.28 &   1.13 &   1.32 \\
                    
\bottomrule
\end{tabular}
\label{tab:mean_number_actions_per_policy}
\end{table}
}

For pipes aged 25 years, Agent-G executes more maintenance actions (\(a_t=1\)), similar to SchM. Agent-E opts for no maintenance, aligning more with RM's strategy. Although CBM carries out some maintenance actions, replacement actions predominate, indicating a greater tendency to permit pipe failures, which explains the observed differences in mean policy costs.

For pipes aged 50 years, CMB offers the most cost-effective policy, with Agent-G's following. CMB conducts fewer maintenance actions and more replacements than Agent-G, accounting for the cost disparity. The policies of Agent-E, RM, and SchM have similar costs. Despite SchM conducting more maintenance, its high number of replacements suggests the maintenance interval requires adjustment. These results indicate that the strategies of CBM, SchM, and RM are less efficient for older pipes due to their higher failure probability.

Regarding the \textit{mean pipe severity level} to assess the impact of various policies on pipe degradation, as shown in Table~\ref{tab:mean_pipe_state_per_policy}.  Our analysis reveals a notable correlation between the average actions per policy, detailed in Table~\ref{tab:mean_number_actions_per_policy}, and the mean pipe severity level. Specifically, the Agent-G control strategy tends to maintain pipes within a severity level of \(k \in [1,2,3]\), whereas the Agent-E, CBM, SchM, and RM policies often result in higher severity levels \(k \in [4,5,F]\), which correlates with increased policy costs. 

{
\setlength{\tabcolsep}{3.8pt} 
\begin{table}[!h]
\centering
\small
\caption{Percentage of severity level per policy obtained with Agent-E, Agent-G, CBM, SchM, and RM, evaluated over 100 episodes in the test environment, for different pipe ages.}
\begin{tabular}{cclllll}
\toprule
Pipe age & Severity &  Agent-E &  Agent-G &  CBM  & SchM &  RM\\
\midrule
\multirow{6}{*}{0} & \(k=1\)  &   59.77 &    58.75 &  59.94 &  59.84 &  58.88 \\
                   & \(k=2\)  &  33.27 &    39.14 &  32.67 &  38.05 &  33.15 \\
                   & \(k=3\)  &  5.39 &     1.70 &   6.00 &   1.79 &   6.36 \\
                   & \(k=4\)  &  1.38 &     0.28 &   1.13 &   0.26 &   1.30 \\
                   & \(k=5\)  &  0.18 &     0.13 &   0.25 &   0.04 &   0.31 \\
                   & \(k=F\)  &  0.01 &     0.01 &   0.01 &   0.01 &   0.01 \\                   
\addlinespace
\multirow{6}{*}{25} & \(k=1\)  &  50.49 &    41.72 &  46.88 &  39.07 &  46.62 \\
                   & \(k=2\)  &  38.96 &    55.27 &  43.09 &  55.55 &  40.86 \\
                   & \(k=3\)  &   8.37 &     2.63 &   8.48 &   4.85 &   9.80 \\
                   & \(k=4\)  &  1.37 &     0.29 &   1.18 &   0.41 &   1.51 \\
                   & \(k=5\)  &  0.78 &     0.07 &   0.36 &   0.10 &   1.18 \\
                   & \(k=F\)  &  0.02 &     0.01 &   0.02 &   0.01 &   0.03 \\
\addlinespace
\multirow{6}{*}{50} & \(k=1\)  & 57.93 &    44.65 &  55.01 &  40.92 &  54.36 \\
                   & \(k=2\)  &  32.58 &    51.40 &  36.14 &  50.46 &  33.09 \\
                   & \(k=3\)  &  7.50 &     3.29 &   7.20 &   7.34 &   9.32 \\
                   & \(k=4\)  &  1.31 &     0.39 &   1.19 &   0.59 &   1.64 \\
                   & \(k=5\)  & 0.65 &     0.25 &   0.43 &   0.67 &   1.55 \\
                   & \(k=F\)  &  0.03 &     0.02 &   0.02 &   0.03 &   0.03 \\                   
\bottomrule
\end{tabular}
\label{tab:mean_pipe_state_per_policy}
\end{table}
}

To summarize, our findings indicate that the Agent-G's policy, derived using \acrshort{DRL}, implements a dynamic management strategy that varies with the pipe's age. This strategy encompasses a more passive approach with new pipes, transitioning to active intervention as the pipes age. This indicates the agent's preference for more frequent maintenance actions rather than allowing pipe failures, which incur higher penalties and replacement costs.

Moreover, Agent-G outperforms Agent-E, illustrating the impact of the degradation model assumption. Specifically, Agent-G's prognostic model \chl{used during training} aligns more closely with the test environment's degradation pattern than Agent-E's, potentially explaining why Agent-G is better equipped to navigate and understand the degradation pattern. This, in turn, enables it to devise a more effective maintenance policy by leveraging a more accurate degradation model.
\subsection{Policy analysis over episode}\label{sec:policy_analysis_over_episode}

\begin{figure*}[!h]
    \centering
    \captionsetup[subfigure]{skip=-7pt}
    \begin{subfigure}[b]{1\linewidth}
        \centering
        \includegraphics[width=1\linewidth]{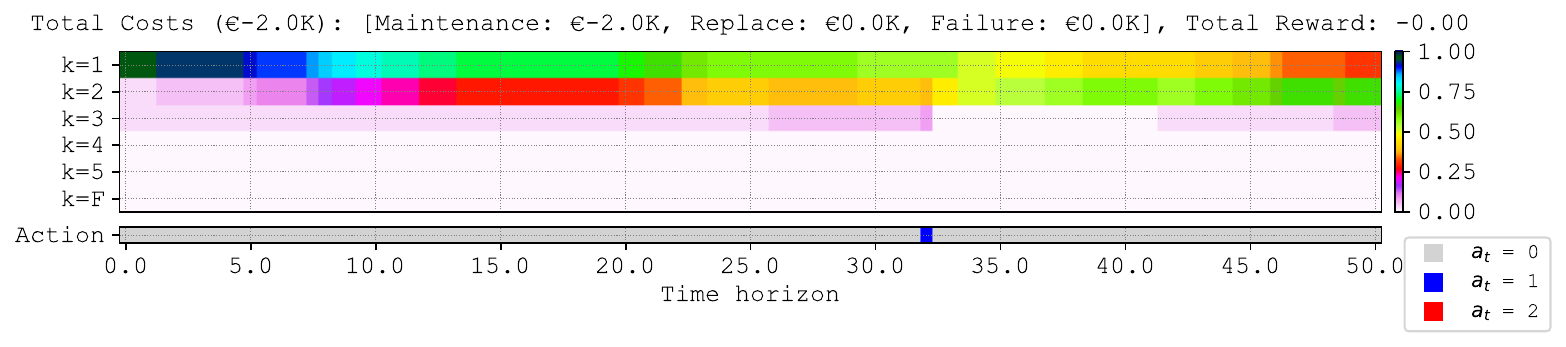}
        \caption{Agent-G}
    \end{subfigure}%
    \\
    \captionsetup[subfigure]{skip=-7pt}
    \begin{subfigure}[b]{1\linewidth}
        \centering
        \includegraphics[width=1\linewidth]{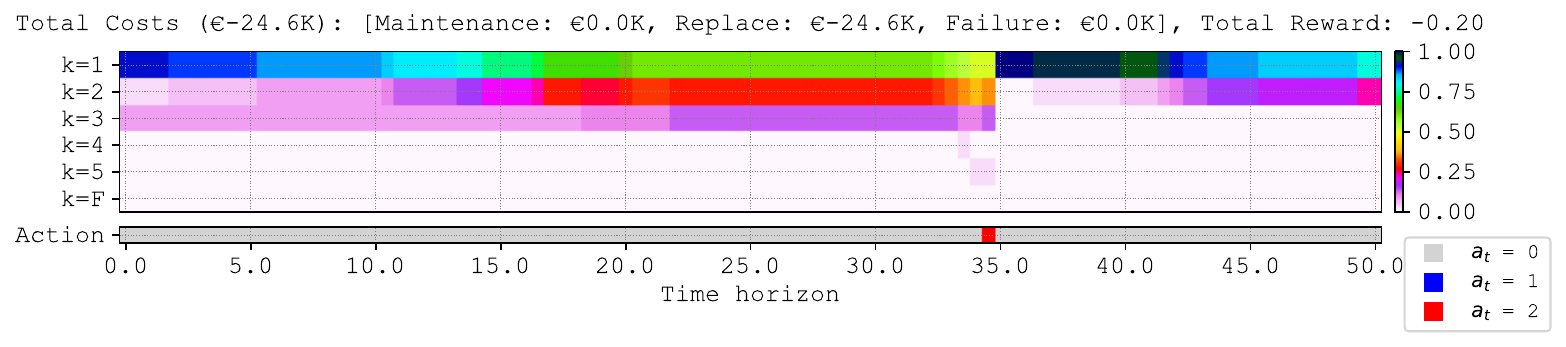}
        \caption{Agent-E}
    \end{subfigure}
    \\
    \captionsetup[subfigure]{skip=-7pt}
    \begin{subfigure}[b]{1\linewidth}
        \centering
        \includegraphics[width=1\linewidth]{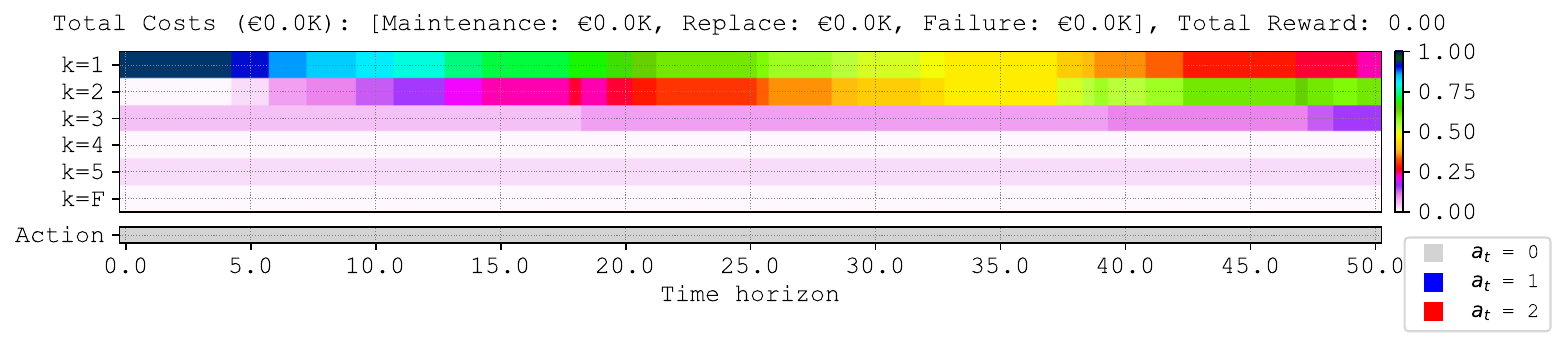}
        \caption{Condition-based Maintenance (CBM)}
    \end{subfigure}
    \\
    \captionsetup[subfigure]{skip=-7pt}
    \begin{subfigure}[b]{1\linewidth}
        \centering
        \includegraphics[width=1\linewidth]{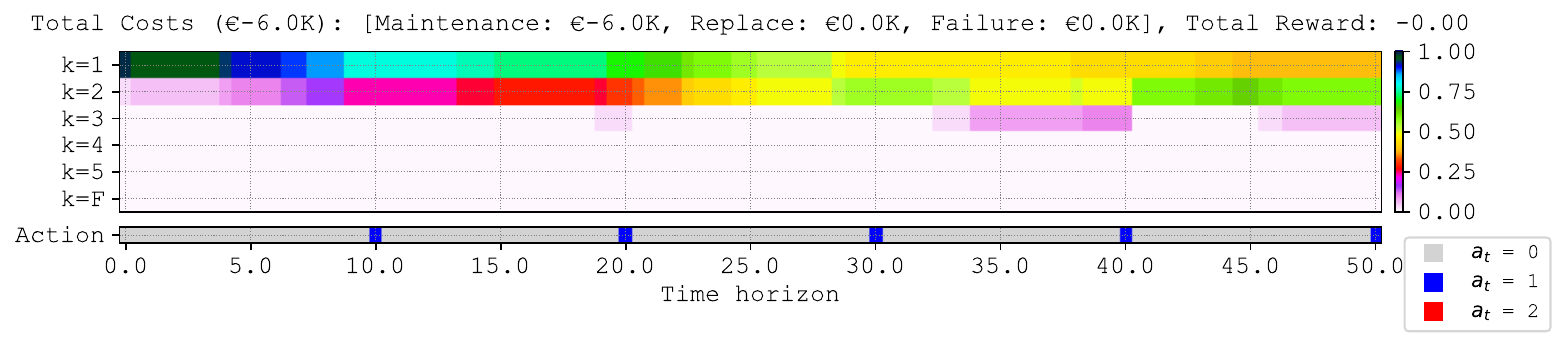}
        \caption{Scheduled Maintenance (SchM)}
    \end{subfigure}
    \caption{Behavior of policies over an episode for a \textbf{new pipe}, showing the health vector over the pipe age and actions per policy: (a) Agent-G, (b) Agent-E, (c) Condition-based Maintenance (CBM), and (d) Scheduled Maintenance (SchM).}
    \label{fig:detailed_policy_view_pipe_age_0}
    \vspace{-10pt}
\end{figure*}

\begin{figure*}[!h]
    \centering
    \captionsetup[subfigure]{skip=-7pt}
    \begin{subfigure}[b]{1\linewidth}
        \centering
        \includegraphics[width=\linewidth]{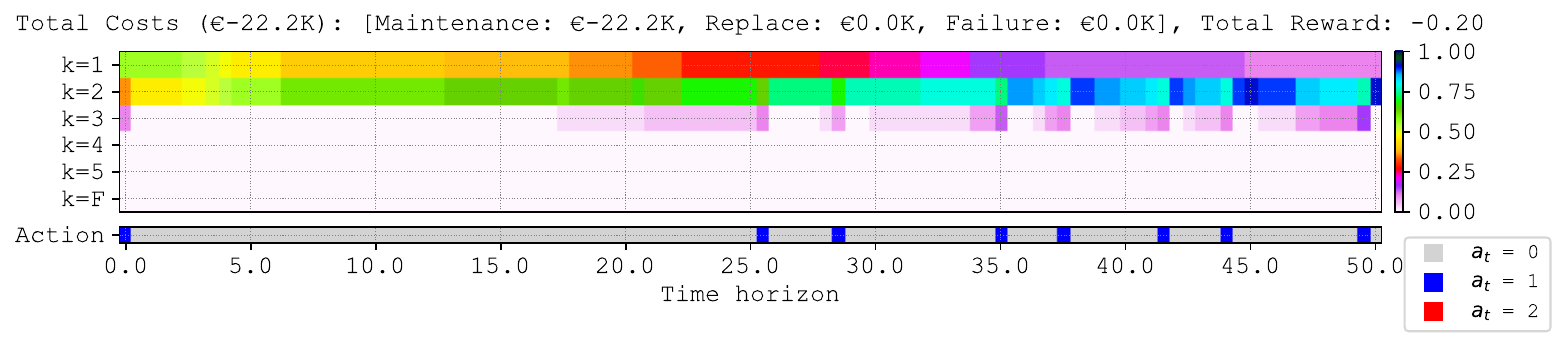}
        \caption{Agent-G}
    \end{subfigure}%
    \\
    \begin{subfigure}[b]{1\linewidth}
        \centering
        \includegraphics[width=\linewidth]{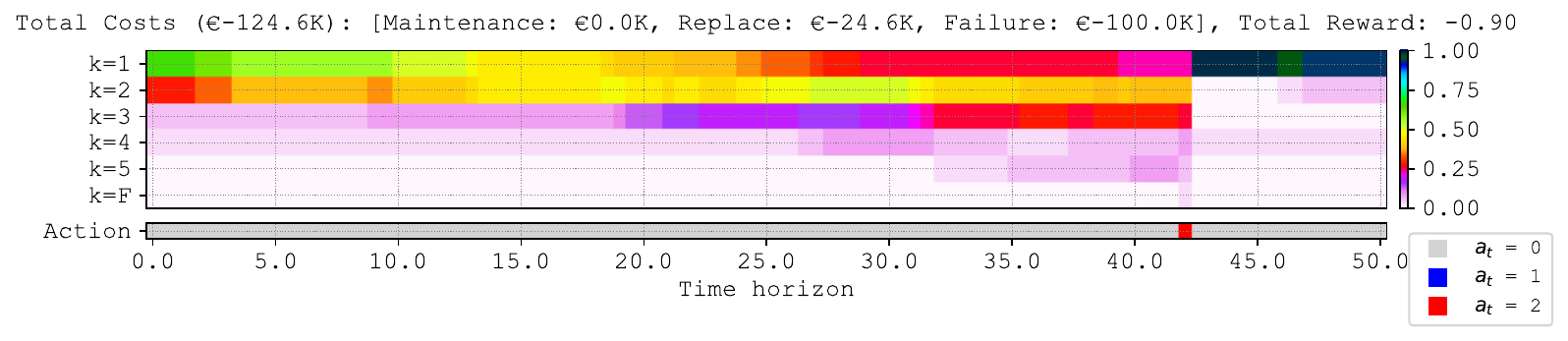}
        \caption{Agent-E}
    \end{subfigure}%
    \\
    \captionsetup[subfigure]{skip=-7pt}
    \begin{subfigure}[b]{1\linewidth}
        \centering
        \includegraphics[width=1\linewidth]{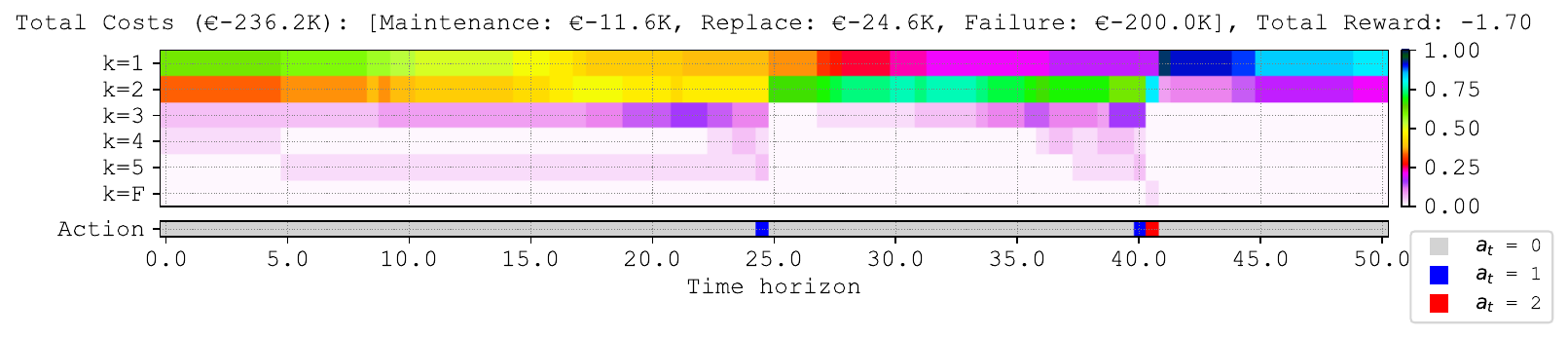}
        \caption{Condition-based Maintenance (CBM)}
    \end{subfigure}
    \\
    \captionsetup[subfigure]{skip=-7pt}
    \begin{subfigure}[b]{1\linewidth}
        \centering
        \includegraphics[width=1\linewidth]{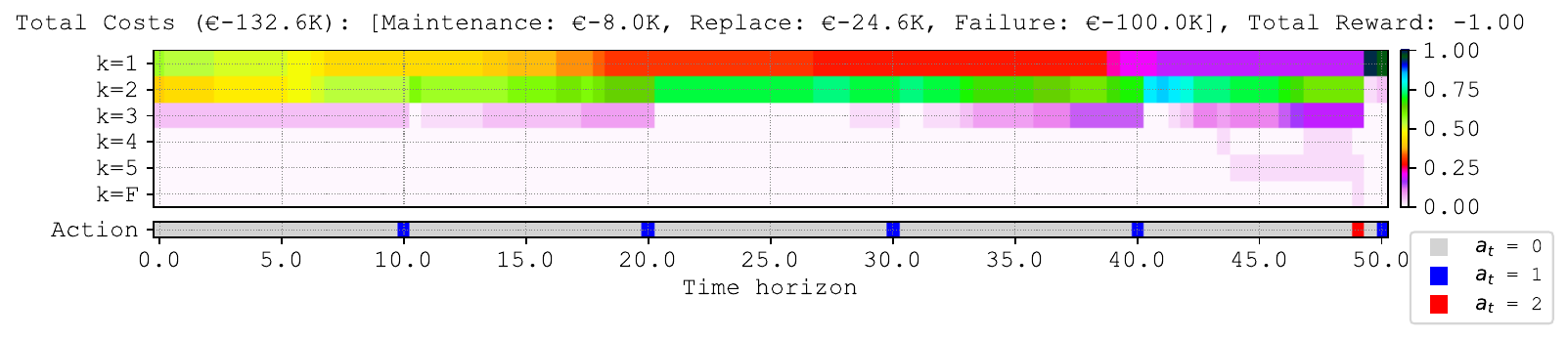}
        \caption{Scheduled Maintenance (SchM)}
    \end{subfigure}
    \caption{Behavior of policies over an episode for a \textbf{pipe aged 25}, showing the health vector over the pipe age and actions per policy: (a) Agent-G, (b) Agent-E, (c) Condition-based Maintenance (CBM), and (d) Scheduled Maintenance (SchM).}
    \label{fig:detailed_policy_view_pipe_age_25}
\end{figure*}

\begin{figure*}[!h]
    \centering
    \captionsetup[subfigure]{skip=-7pt}
    \begin{subfigure}[b]{1\linewidth}
        \centering
        \includegraphics[width=1\linewidth]{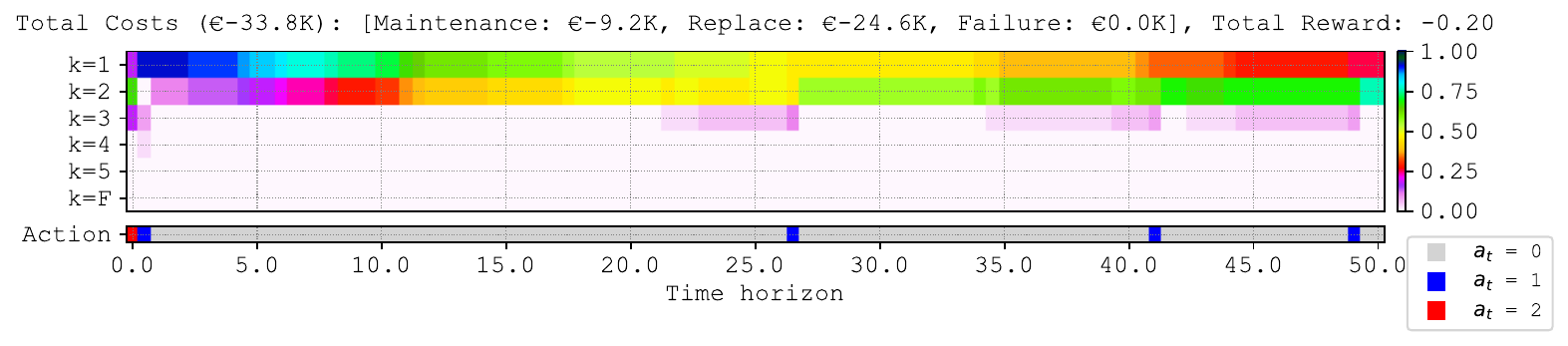}
        \caption{Agent-G}
    \end{subfigure}%
    \\
    \captionsetup[subfigure]{skip=-7pt}
    \begin{subfigure}[b]{1\linewidth}
        \centering
        \includegraphics[width=1\linewidth]{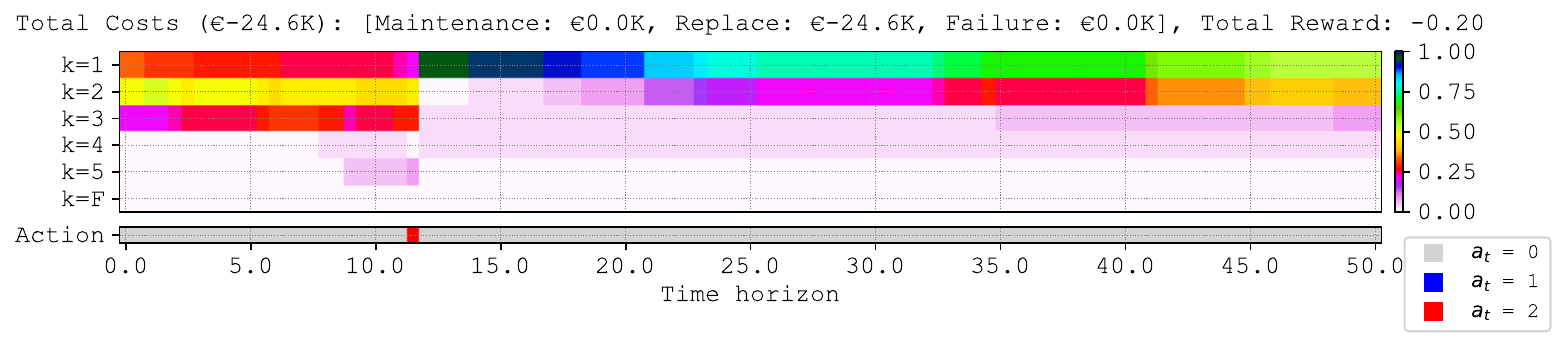}
        \caption{Agent-E}
    \end{subfigure}%
    \\
    \captionsetup[subfigure]{skip=-7pt}
    \begin{subfigure}[b]{1\linewidth}
        \centering
        \includegraphics[width=1\linewidth]{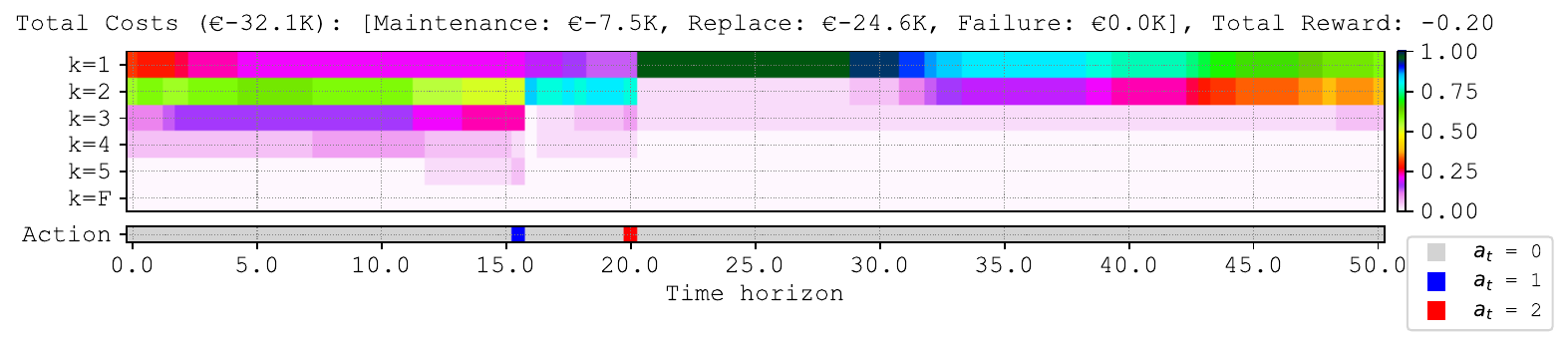}
        \caption{Condition-based Maintenance (CBM)}
    \end{subfigure}
    \\
    \captionsetup[subfigure]{skip=-7pt}
    \begin{subfigure}[b]{1\linewidth}
        \centering
        \includegraphics[width=1\linewidth]{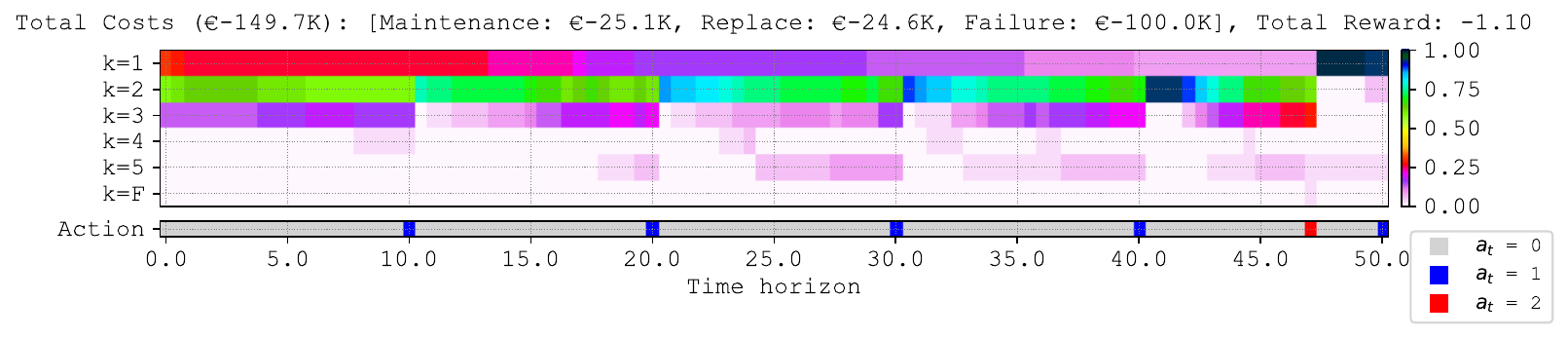}
        \caption{Scheduled Maintenance (SchM)}
    \end{subfigure}
    \caption{Behavior of policies over an episode for a \textbf{pipe aged 50}, showing the health vector over the pipe age and actions per policy: (a) Agent-G, (b) Agent-E, (c) Condition-based Maintenance (CBM), and (d) Scheduled Maintenance (SchM).}
    \label{fig:detailed_policy_view_pipe_age_50}
\end{figure*}

In Section \ref{sec:policy_analysis_overview}, we present an overview of policy performances. This section delves into the details per episode to provide further understanding on these policies. Figures \ref{fig:detailed_policy_view_pipe_age_0}, \ref{fig:detailed_policy_view_pipe_age_25}, and \ref{fig:detailed_policy_view_pipe_age_50} detail the performance of the Agent-E, Agent-G, CMB, and SchM policies for pipes with ages 0, 25 and 50, respectively. The RM heuristic is excluded from this analysis due to its straightforward approach: allowing the pipe to fail before replacing it.

Figure \ref{fig:detailed_policy_view_pipe_age_0} shows that for a brand new pipe: (a) Agent-G performs maintenance on the pipe at approximately 32 years old; (b) Agent-E opts to replace the pipe when it is around 35 years old, which may be attributed to the presence of elements with higher severity levels in that specific episode; (c) CBM chooses not to act, which results in the least expensive policy in this comparison. However, it is observed that some pipe sections reach severity level $k=5$ throughout the episode. Not taking any action is deemed risky since progressing to $k=F$ becomes more likely and incurs higher costs; (d) \chl{SchM effectively controls severity levels but is more expensive than Agent-G's policy due to more frequent maintenance actions.}

Figure \ref{fig:detailed_policy_view_pipe_age_25} shows that for a pipe aged 25: (a) Agent-G exhibits increased activity, indicating more frequent maintenance actions, especially as the pipe ages to 50, shortening the maintenance intervals; (b) Agent-E postpones any action until the pipe fails, at which point it replaces the pipe with a new one, akin to RM; (c) CBM also initiates maintenance around the pipe's 50-year mark. However, degradation escalates from age 60, leading to failure at 66. The inability to manage this increased severity results in significant penalty costs, diminishing the effectiveness of this policy; (d) Similarly, SchM manages severity levels effectively until the pipe reaches approximately 70 years of age, at which point degradation accelerates, resulting in failure at 73.

Figure \ref{fig:detailed_policy_view_pipe_age_50} shows that for a pipe aged 50: (a) Agent-G opts to replace the pipe at age 50, followed by maintenance in the subsequent time step. This decision is likely influenced by parts of the pipe being at severity levels $k \in {3,4}$. Such a scenario is plausible, as new pipes can exhibit high severity levels at a young age due to defects in the material or errors during the construction and installation process. This concept is represented in the MSDM by the initial probability state vector ($S_{k}^0$). Additionally, Agent-G recommends maintenance at the interval when the pipe reaches the age of 26 years; (b) Agent-E suggests replacement at approximately 62 years, without recommending further maintenance; (c) CMB advocates for maintenance at about 65 years, followed by replacement at 70 years, in line with heuristics described in Section \ref{sec:comparison_heuristics}; (d) SchM consistently performs maintenance at regular intervals, yet faces significant degradation, culminating in failure around 97 years.

\section{Discussion and Conclusions}\label{sec:discussion_conclusions}

In this paper, we explore the applications of Prognostics and Health Management (PHM) in sewer pipe asset management. Our study focuses on component-level (i.e., pipe-level) maintenance policy optimization by integrating stochastic multi-state degradation modeling and Deep Reinforcement Learning (DRL). The goal is to assess the effectiveness of DRL in deriving cost-effective maintenance strategies tailored to the specific conditions and requirements of sewer pipes.

A key contribution of our work is the integration of prognostics models with a maintenance policy optimization framework. We utilize a tailored reward function that aligns with damage severity levels, enabling a more complex and realistic maintenance optimization setup.

Our methodology includes a real-world case study from a Dutch sewer network, which provides historical inspection data. Through hyper-parameter tuning and policy analysis, we benchmark our optimized policies against traditional heuristics, including condition-based, scheduled, and reactive maintenance.

Our findings suggest that agents trained with the Proximal Policy Optimization algorithm are highly capable of developing strategic maintenance policies, adapting to pipe age, and surpassing heuristic baselines by learning cost-effective dynamic management strategies.

To evaluate the impact of degradation model assumptions, we trained one agent using the Gompertz probability density function and another using the Exponential probability density function.

During testing, both agents were assessed in an environment parameterized with the Weibull probability density function. The Gompertz-trained agent, whose behavior more closely resembled the Weibull model, demonstrated better generalization, resulting in more effective maintenance policies compared to the Exponential-trained agent.

\vspace{-15pt}
\paragraph{Future work:} The following directions are identified:
\begin{compactitem}
    \item Advancing toward partially observable state spaces with the introduction of inspection actions, considering context, and leveraging deep learning capabilities.
    \item Utilizing knowledge acquired by agents to develop explainable and robust heuristics.
    \item Although this paper focused on a single cohort of pipes, studies in \cite{jimenez2022deterioration,jimenez2024comparison} show different cohorts exhibit varied dynamics, highlighting the importance of understanding how RL agents adapt.
    \item Comparing RL-based approaches with other policy optimization algorithms to better understand the capacity of RL methods to achieve global-optima maintenance strategies.
    \item Investigating various reward functions \chl{(e.g., dense)} and RL algorithms to determine the most effective for devising maintenance policies.
    \item \chl{Extent to system-level analysis and evaluate aspects such as scalability.}
    \item Moving toward multi-infrastructure asset management to promote coordinated management for optimizing costs and minimizing disruption from interventions.
\end{compactitem}

\section*{Acknowledgements}
{\small
This research has been partially funded by NWO under the grant PrimaVera (https://primavera-project.com) number NWA.1160.18.238.
}

\bibliographystyle{apacite}
\PHMbibliography{ijphm}

\section*{Biographies}

\noindent\textbf{Lisandro A. Jimenez-Roa} is a doctoral candidate in Computer Science at the University of Twente, The Netherlands. He has a background in civil engineering and has contributed to various projects in structural health monitoring, finite element modeling, and damage detection through data analytics and machine learning. His current research focuses on Prognostics and Health Management, specifically engineering systems within the PrimaVera project (\textbf{https://primavera-project.com}), emphasizing multi-state stochastic degradation modeling and maintenance policy optimization using Reinforcement Learning techniques.

\noindent\textbf{Thiago D. Sim\~{a}o} is an Assistant Professor in the Eindhoven University of Technology, the Netherlands.
He obtained his PhD in Computer Science at Delft University of Technology.
Previously, he was a PostDoc researcher at Radboud University Nijmegen.
His research interests lie primarily in reliably automating sequential decision-making, focusing on reinforcement learning.

\noindent\textbf{Zaharah Bukhsh} is an assistant professor at Eindhoven University of Technology, Eindhoven, Netherlands . She holds a Master's degree in computer science and a Ph.D. in engineering technology from University of Twente, Enschede, Netherlands. Her research focuses on developing data-driven methods with deep learning and deep reinforcement learning. Her research targets broad application areas including asset management, scheduling, and resource optimization. She has contributed to several H2020 and NWO research projects.

\noindent\textbf{Tiedo Tinga} is a full professor in dynamics based maintenance at the University of Twente since 2012 and full professor of Life Cycle Management at the Netherlands Defence
Academy since 2016. He received his Ph.D. degree in mechanics of materials from Eindhoven University in 2009. He is chairing the smart maintenance knowledge center and
leads a number of research projects on developing predictive maintenance concepts, mainly based on the physics of failure models, but also following data-driven approaches.

\noindent\textbf{Hajo Molegraaf} completed his PhD at the University of Groningen and has worked as an assistant and postdoc researcher at the University of Geneva and Yale University. Since October 2022, Molegraaf joined as a Research Fellow within the Formal Methods and Tools (FMT) group in the EEMCS faculty at Twente. Additionally, Molegraaf is a co-founder and software developer at Rolsch Assetmanagement, a company based in Enschede, The Netherlands.

\noindent\textbf{Nils Jansen} is a full professor at the Ruhr-University Bochum, Germany, and leads the chair of Artificial Intelligence and Formal Methods. The mission of his chair is to increase the trustworthiness of Artificial Intelligence (AI). Prof. Jansen is also an associate professor at Radboud University, Nijmegen, The Netherlands. He was a research associate at the University of Texas at Austin and received his Ph.D. with distinction from RWTH Aachen University, Germany. His research is on intelligent decision-making under uncertainty, focusing on formal reasoning about the safety and dependability of artificial intelligence (AI). He holds several grants in academic and industrial settings, including an ERC starting grant titled Data-Driven Verification and Learning Under Uncertainty (DEUCE). 

\noindent\textbf{Mari\"{e}lle~Stoelinga} is a full professor of risk analysis for high-tech systems, both at the University of Twente and Radboud University, the Netherlands. She holds a Master's degree in Mathematics \& Computer Science, and a Ph.D. in Computer Science. After her Ph.D., she has been a postdoctoral researcher at the University of California at Santa Cruz, USA. 
Prof. Stoelinga leads various research projects, including a large national consortium on Predictive Maintenance and an ERC consolidator grant on safety and security interactions.

\appendix
\renewcommand{\thesection}{Appendix \Alph{section}}

\section{Parameters of multi-state degradation models}\label{app:msdm_parameters}


\begin{table}[!h]
\centering
\small
\caption{\acrshort{MSDM} hyper-parameters for cohort \texttt{CMW}, using hazard functions modeled with the \textit{exponential} (\(\lambda^{E}(t|\epsilon)\)), \textit{Gompertz} (\(\lambda^{G}(t|\alpha,\beta)\)), and \textit{Weibull} (\(\lambda^{W}(t|\eta,\rho)\)) probability density functions.}
\begin{tabular}{@{}rccccc@{}}
& $\lambda^E(t|\epsilon)$ & \multicolumn{2}{c}{$\lambda^G(t|\alpha,\beta)$} & \multicolumn{2}{c}{$\lambda^W(t|\eta,\rho)$}
\\ \midrule
$i \to j$ & $\epsilon$  & $\alpha$ & $\beta$ & $\eta$ & $\rho$ \\ \midrule
$1 \to 2$ &  2.4E-02 & 2.3E+00 & 8.4E-03 & 1.3E+00 & 4.4E+01 \\
$2 \to 3$ &  9.4E-03 & 2.1E-02 & 5.5E-02 & 2.9E+00 & 7.7E+01 \\
$3 \to 4$ &  5.7E-03 & 3.3E+00 & 2.8E-03 & 3.5E+00 & 8.1E+01 \\
$4 \to 5$ &  1.8E-02 & 2.4E+00 & 8.7E-03 & 7.0E+00 & 5.5E+01 \\
$1 \to F$ &  3.0E-18 & 1.4E-01 & 3.1E-04 & 4.1E-06 & 4.6E+01 \\
$2 \to F$ &  6.0E-04 & 8.8E-01 & 7.0E-19 & 2.7E-04 & 4.6E+01 \\
$3 \to F$ &  1.0E-18 & 2.2E-03 & 4.5E-02 & 3.0E-05 & 4.7E+01 \\
$4 \to F$ &  1.0E-18 & 9.8E-05 & 8.6E-03 & 1.1E-03 & 4.5E+01 \\
$5 \to F$ &  1.0E-18 & 7.0E-19 & 3.8E-01 & 1.7E+00 & 5.9E+01 \\
\bottomrule
\end{tabular}
\label{tab:hyperparameter_msdm_CMW}
\end{table}

\begin{table}[]
\centering
\small
\caption{Initial state vector \(S_k^0\) for MSDM of cohort \texttt{CMW}.}
\begin{tabular}{cccc}
\hline
\(S^0_k\) & Exponential & Gompertz & Weibull \\ \hline
$k=1$   & 9.89E-01 & 9.58E-01 & 9.23E-01 \\
$k=2$   & 1.26E-17 & 0.00E+00 & 2.59E-02 \\
$k=3$   & 3.70E-23 & 4.00E-02 & 3.10E-02 \\
$k=4$   & 1.11E-02 & 1.61E-03 & 1.13E-02 \\
$k=5$   & 2.11E-22 & 2.00E-15 & 2.07E-03 \\
$k=F$   & 3.87E-22 & 1.56E-04 & 6.40E-03 \\
\bottomrule
\end{tabular}
\label{tab:initial_state_vector_CWM}
\end{table}

\end{document}